\pdfoutput=1
\documentclass[11pt]{article}

\usepackage[preprint]{acl}

\usepackage{times}
\usepackage{latexsym}
\usepackage{xcolor}
\definecolor{pending}{gray}{0.65}
\usepackage{amsmath}
\usepackage{amssymb}
\usepackage{amsfonts}
\usepackage{booktabs}
\usepackage{placeins}
\usepackage[T1]{fontenc}
\usepackage[utf8]{inputenc}

\usepackage{microtype}

\usepackage{inconsolata}
\usepackage{float}
\usepackage{graphicx}
\usepackage{subcaption}
\title{Probing Ethical Framework Representations in Large Language Models: Structure, Entanglement, and Methodological Challenges}

\author{Weilun Xu$^{\dagger}$ \quad Alexander Rusnak$^{\ddagger}$ \quad Fr\'{e}d\'{e}ric Kaplan$^{\ddagger}$ \\
  $^{\dagger}$School of Computer and Communication Sciences, \'{E}cole Polytechnique F\'{e}d\'{e}rale de Lausanne \\
  $^{\ddagger}$Digital Humanities Lab, \'{E}cole Polytechnique F\'{e}d\'{e}rale de Lausanne \\
  \texttt{\{weilun.xu, alexander.rusnak, frederic.kaplan\}@epfl.ch}}

\begin{document}
\maketitle
\begin{abstract}
When large language models make ethical judgments, do their internal representations distinguish between normative frameworks, or collapse ethics into a single acceptability dimension? We probe hidden representations across five ethical frameworks (deontology, utilitarianism, virtue, justice, commonsense) in six LLMs spanning 4B--72B parameters. Our analysis reveals differentiated ethical subspaces with asymmetric transfer patterns---e.g., deontology probes partially generalize to virtue scenarios while commonsense probes fail catastrophically on justice. Disagreement between deontological and utilitarian probes correlates with higher behavioral entropy across architectures, though this relationship may partly reflect shared sensitivity to scenario difficulty. Post-hoc validation reveals that probes partially depend on surface features of benchmark templates, motivating cautious interpretation. We discuss both the structural insights these methods provide and their epistemological limitations.
\end{abstract}

\section{Introduction}

\begin{figure}[t] \centering \includegraphics[width=\columnwidth]{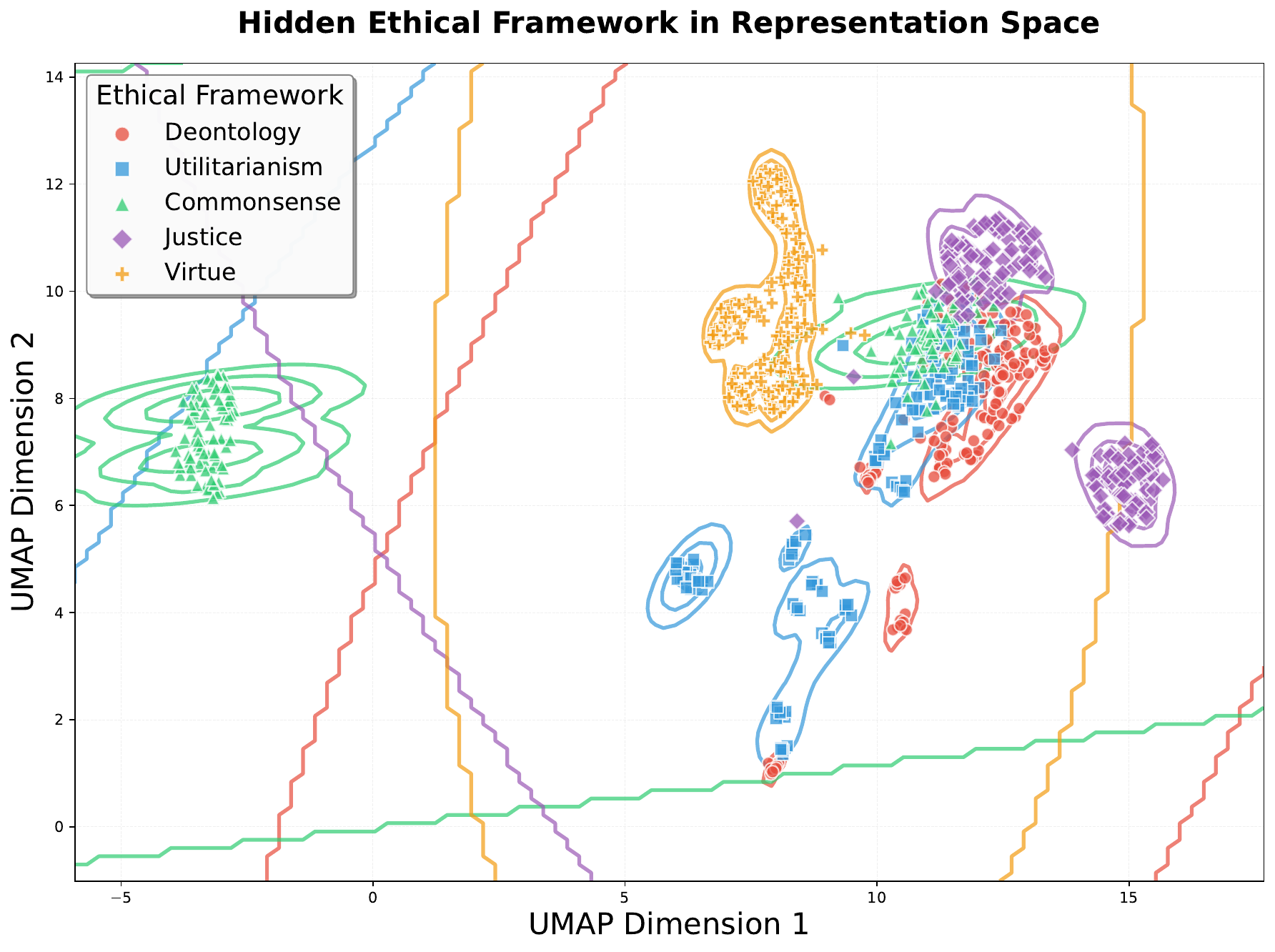} \caption{UMAP projection of activation patterns at layer 20 in Mistral-7B-Instruct. This visualization illustrates the differentiated yet entangled nature of ethical representations---partial overlap among action-oriented frameworks (red, blue, green) versus separation for character-focused virtue ethics (orange). Note that UMAP projections are for qualitative illustration only; quantitative claims about framework relationships rest on the cross-framework transfer matrices (Figure~\ref{fig:cross_framework}) and probe accuracy metrics, not visual clustering patterns. The structural entanglement suggested here is validated formally across scales (4B--72B) in Appendix~\ref{app:entanglement}.} \label{fig:conceptual} \end{figure}

Models tuned to produce ethically appropriate outputs can appear aligned while relying on shallow pattern-matching of training distributions \cite{hendrycks2021aligning}. Moral biases emerge implicitly from pretraining alone \cite{simmons2022moral, ziems2022moral}, but whether models organize these biases into distinct normative subspaces or compress them into a single ``good/bad'' dimension is unknown. The distinction matters: a model with genuinely differentiated ethical representations might fail in predictable, diagnosable ways, while one that merely memorizes acceptability labels could fail silently. We investigate this question by mapping the internal geometry of ethical representations across architectures and scales.

We probe hidden representations across five ethical frameworks to investigate this question at the representational level. A persistent challenge in probing methodology is the interpretation gap: high classification accuracy may reflect genuine concept encoding or merely surface correlates \cite{belinkov2022probing, hewitt2019designing}. We navigate these interpretive challenges by grounding our probing analysis in measurable behavioral outcomes. Rather than debating whether these patterns constitute functional moral reasoning \cite{mahowald2024dissociating}, we focus on their potential behavioral relevance. By examining whether probe-identified patterns correlate with output stability, we investigate whether this internal structure is behaviorally relevant, while acknowledging significant methodological caveats.

Our analysis reveals three structural properties that function as a chain of evidence for model reliability. First, representations are differentiated but entangled. Asymmetric transfer patterns (e.g., Deontology $\rightarrow$ Virtue succeeds; Commonsense $\rightarrow$ Justice fails) rule out a collapsed single-dimension encoding (Figure~\ref{fig:conceptual}), indicating models maintain distinct, non-interchangeable ethical subspaces. Second, this differentiation is imperfect: probes maintain high confidence even during transfer failures, indicating ethical scenarios simultaneously activate multiple framework-specific patterns without the structure needed to distinguish appropriate from inappropriate application.

Third, internal conflict correlates with external instability. When deontological and utilitarian probes make confident opposing predictions, models exhibit significantly higher choice entropy. This conflict-entropy correlation, while replicated across architectures (7B--72B parameters), requires further validation to rule out confounds such as scenario difficulty. These findings demonstrate that probing can reveal meaningful structural properties of ethical representations, while highlighting limitations that must be addressed before such methods serve as reliable diagnostics.

\section{Related Work}

\subsection{Probing Internal Representations}
Probing techniques reveal how neural networks encode information, ranging from basic linguistic features \cite{belinkov2017neural, tenney2019bert} to latent semantic dimensions discovered through unsupervised methods like CCS and CCR \cite{burns2023discovering, stoehr2024unsupervised}. However, a persistent critique challenges this methodology: high probe accuracy may reflect surface correlates or memorization rather than functional concept encoding \cite{hewitt2019designing, belinkov2022probing}.

We bridge this interpretive gap by linking linear separability to causal utility. Recent findings demonstrate that linear probes often identify ``representation engineering'' directions that causally govern model outputs, from truthfulness \cite{zou2023representation, li2024inference} to belief states \cite{kosinski2023theory, rimsky2023steering}. Motivated by this evidence that linear features frequently support causal control, we examine whether our ethical probes' internal conflicts correlate with behavioral entropy. This suggests that the structures we identify may be behaviorally relevant, though confounds have not been fully controlled.

Critical evaluation of probing methodology has refined interpretation standards. \citet{hewitt2019designing} introduced control tasks demonstrating that probe accuracy alone cannot distinguish genuine encoding from memorization, while \citet{belinkov2022probing} catalogued systematic limitations. We incorporate these insights by anchoring our probing metrics to observable behavioral outcomes, focusing on the functional consequence of representational structures rather than their semantic ontology.

This diagnostic focus distinguishes our work from activation steering \cite{tlaie2024exploring, choi2024moral}, which actively modifies behavior and may confound training effects with native inclinations. Probing preserves model integrity, allowing us to map native representational geometry alongside potential structural failures.

\paragraph{Internal Consistency and Knowledge Conflicts.}
Our conflict-entropy approach connects to broader work on predicting behavioral reliability from internal dynamics. Semantic entropy methods detect hallucinations by measuring consistency across sampled outputs \cite{kuhn2023semantic, farquhar2024detecting}, while self-consistency approaches aggregate multiple reasoning paths to improve factual accuracy \cite{wang2023selfconsistency}. These methods operate at the output level; we extend this logic to the representational level, detecting conflicts before generation. 

More directly related is the knowledge conflict literature examining tensions between parametric knowledge (PK) and contextual knowledge (CK). \citet{longpre2021entity} and \citet{neeman2023disentqa} demonstrate that models exhibit predictable behavioral patterns when internal knowledge conflicts with provided context, with entropy and logit dynamics serving as conflict indicators. Our ethical framework conflicts represent an analogous phenomenon: competing representational signals that manifest as behavioral switching. The key difference is that ethical conflicts occur entirely within parametric representations---between framework-specific encodings learned during pretraining---rather than between parametric and contextual sources.

\paragraph{Probe-Based Risk Detection.}
Our diagnostic approach builds on work using probes to detect reliability risks before they manifest in outputs. Truthfulness probes \cite{azaria2023internal, burns2023discovering} identify when models ``know'' their outputs are false, while hallucination detection methods use internal states to flag unreliable generations \cite{kadavath2022language, li2024inference}. These approaches typically probe for a single binary property (true/false, hallucinated/grounded). We extend this paradigm to multi-dimensional ethical space, where risk emerges not from a single unreliability signal but from conflict between multiple framework-specific representations. This positions our conflict score as analogous to existing probe-based risk detectors, but targeting a distinct failure mode: inconsistency arising from competing ethical signals rather than factual uncertainty.

\paragraph{Formal versus Functional Competence.}
We ground our analysis in the distinction between formal linguistic competence (statistical knowledge) and functional competence (reasoning) \cite{mahowald2024dissociating}. This competence-performance dissociation, well-established in cognitive neuroscience \cite{fedorenko2024language}, offers a precise mechanism for AI alignment failures: models may essentially ``memorize'' ethical patterns (formal encoding) without integrating them into a coherent decision-making framework (functional consistency). Our finding that internal probe disagreement correlates with behavioral inconsistency is consistent with this hypothesis—it suggests that while models possess sophisticated formal maps of ethical frameworks, these representations can actively conflict during processing, destabilizing the final output. Because this internal conflict correlates with external entropy, our methodology suggests potential auditing relevance independent of theoretical debates regarding machine understanding, though post-hoc validation reveals probes partially rely on surface features (see Limitations), qualifying the strength of this link.

\subsection{Moral Reasoning in Language Models}

Behavioral evaluations have established that LLMs exhibit distinct moral biases, often varying by architecture and scale. Proprietary models tend toward utilitarian reasoning while open-source alternatives display deontological tendencies \cite{scherrer2024evaluating}, with larger models frequently defaulting to rule-based responses even when consequences are negative \cite{yuan2024right}. Notably, these patterns emerge prior to alignment: base models display clear Moral Foundations biases \cite{simmons2022moral} and implicit normative assumptions \cite{ziems2022moral, neuman2025analyzing} derived solely from pretraining distributions.

Yet behavioral consistency does not imply representational coherence. \citet{ganguli2023capacity} showed that while Constitutional AI can steer surface outputs, it leaves internal representations opaque. Similarly, while activation steering can successfully manipulate ethical orientation along specific vectors \cite{tlaie2024exploring, park2024ai}, understanding the broader geometric relationship between competing frameworks requires complementary observational methods.

This leaves an explanatory gap: we know models can mimic ethical reasoning \cite{hendrycks2021aligning, tanmay2023probing}, but lack understanding of the internal mechanisms sustaining it. When models fail—rejecting instrumental harm for impartial beneficence even against human preference \cite{marraffini2024greatest}—behavioral metrics offer little diagnostic power. We address this by shifting focus from what models choose to how their internal representations compete, providing a structural explanation for why ethical judgments become unstable under pressure.

\section{Methods}

\subsection{Ethical Framework Datasets}
To characterize the internal diversity of moral representations, we utilize the ETHICS benchmark \cite{hendrycks2021aligning}, which spans five distinct normative theories: deontology (rules), utilitarianism (consequences), virtue (character), justice (fairness), and commonsense (intuition).

We convert all scenarios into standardized binary-choice prompts. For utilitarianism, we randomize scenario positioning (A vs. B) to eliminate positional bias, creating a robust binary classification task (details in Appendix~\ref{app:prompts}). This multi-framework design is central to our approach: if models merely collapsed ethics into a single ``good/bad'' dimension, probe transfer would be uniform. We specifically test for differentiated transfer patterns to demonstrate the existence of specialized representational subspaces.

\subsection{Probing Methodology}
We employ linear logistic regression probes to decode framework-specific patterns from frozen hidden states. For a representation $h_i^{(l)} \in \mathbb{R}^d$ at layer $l$, the probe learns $P(y=1|h_i^{(l)}) = \sigma(w^T h_i^{(l)} + b)$. Probes are trained independently for each framework, ensuring no cross-contamination of ethical signals.

We intentionally select linear classifiers to prioritize interpretive transparency. If a linear probe can distinguish frameworks, it implies the model has already disentangled these concepts into linearly accessible formats usable for generation. This establishes a ``lower bound'' on the model's structural clarity, ensuring our findings reflect intrinsic model geometry rather than the capacity of a complex non-linear probe.

\paragraph{Validation Strategy.} To distinguish genuine conceptual encoding from surface-level memorization without relying on synthetic control tasks \citep{hewitt2019designing}, we pre-register three validity conditions. First, probes must demonstrate specificity through asymmetric transfer—succeeding on related frameworks while failing on orthogonal ones—rather than the uniform performance expected from memorizing generic surface cues. Second, probe signals must be predictive, meaning internal disagreement should statistically predict behavioral inconsistency (entropy). Third, patterns must be replicable across distinct model families, confirming they capture general transformer properties rather than architecture-specific artifacts. We note that subsequent validation experiments (reported in Limitations) revealed that probes partially depend on surface features, qualifying the strength of these validity conditions. We report our original analysis alongside these post-hoc findings for transparency.

\subsection{Selected LLMs}
We analyze six LLMs selected to span key axes of scale (4B--72B) and training methodology. We present Mistral-7B-Instruct \citep{jiang2023mistral} as our primary case study for conflict analysis due to its interpretable bimodal conflict distribution, and Llama-3.3-70B-Instruct \citep{grattafiori2024llama3} for layer-wise analysis to demonstrate patterns at frontier scale. Additional models—Qwen-2.5 (7B/72B) \citep{yang2024qwen2}, DeepSeek-R1-Distill-7B \citep{guo2025deepseek}, and Gemma-3-4B \citep{team2025gemma3}—ensure our findings generalize across architectural families.

We deliberately avoid aggregating results into a single ``average'' metric. Distinct models exhibit qualitatively different ethical profiles—such as virtue isolation in Qwen or commonsense divergence in DeepSeek—that aggregate statistics would obscure. Because probe-based analyses aim to characterize individual architectural signatures, maintaining model-specific resolution is methodologically essential.

\subsection{Within-Framework Probe Training}
To map the developmental trajectory of ethical representations, we train independent linear probes at every transformer layer. This validates that framework-specific encodings become sufficiently differentiated at intermediate depths to support meaningful conflict detection, and identifies the optimal depth balancing representational maturity with preserved diversity.

\subsection{Cross-Framework Entanglement Analysis}
\label{sec:cross_framework}

We investigate whether ethical frameworks encode orthogonal dimensions or share overlapping features through systematic cross-framework probe transfer. For each framework pair, we train a probe on one framework's training set and evaluate on all others, yielding a $5 \times 5$ transfer matrix.

To isolate transfer effects from layer-specific confounds, we evaluate all probes at a unified layer corresponding to 65\% model depth (e.g., Layer 52 for Llama-3.3-70B). We select this depth based on two empirical criteria: it represents the point where all frameworks achieve sufficient maturity for valid comparison (overcoming early-layer noise), and it marks the peak of cross-framework generalization before models degrade into narrow specialization in final layers (Appendix~\ref{app:entanglement}).

We employ three metrics to characterize these relationships. Accuracy quantifies whether a probe trained on framework $F_{\text{train}}$ can correctly classify scenarios from $F_{\text{test}}$, where high off-diagonal accuracy indicates shared structure. Confidence reveals whether probes recognize out-of-distribution scenarios; high confidence despite poor accuracy signals miscalibration. We quantify this using Expected Calibration Error (ECE), measuring the gap between predicted confidence and actual accuracy. This calibration analysis is essential: if probes remain confident even when wrong, then scenarios with conflicting probe predictions might reflect miscalibration rather than genuine representational conflict.

\begin{figure*}[!t] \centering \begin{subfigure}[t]{0.5\textwidth} \centering \includegraphics[width=\linewidth]{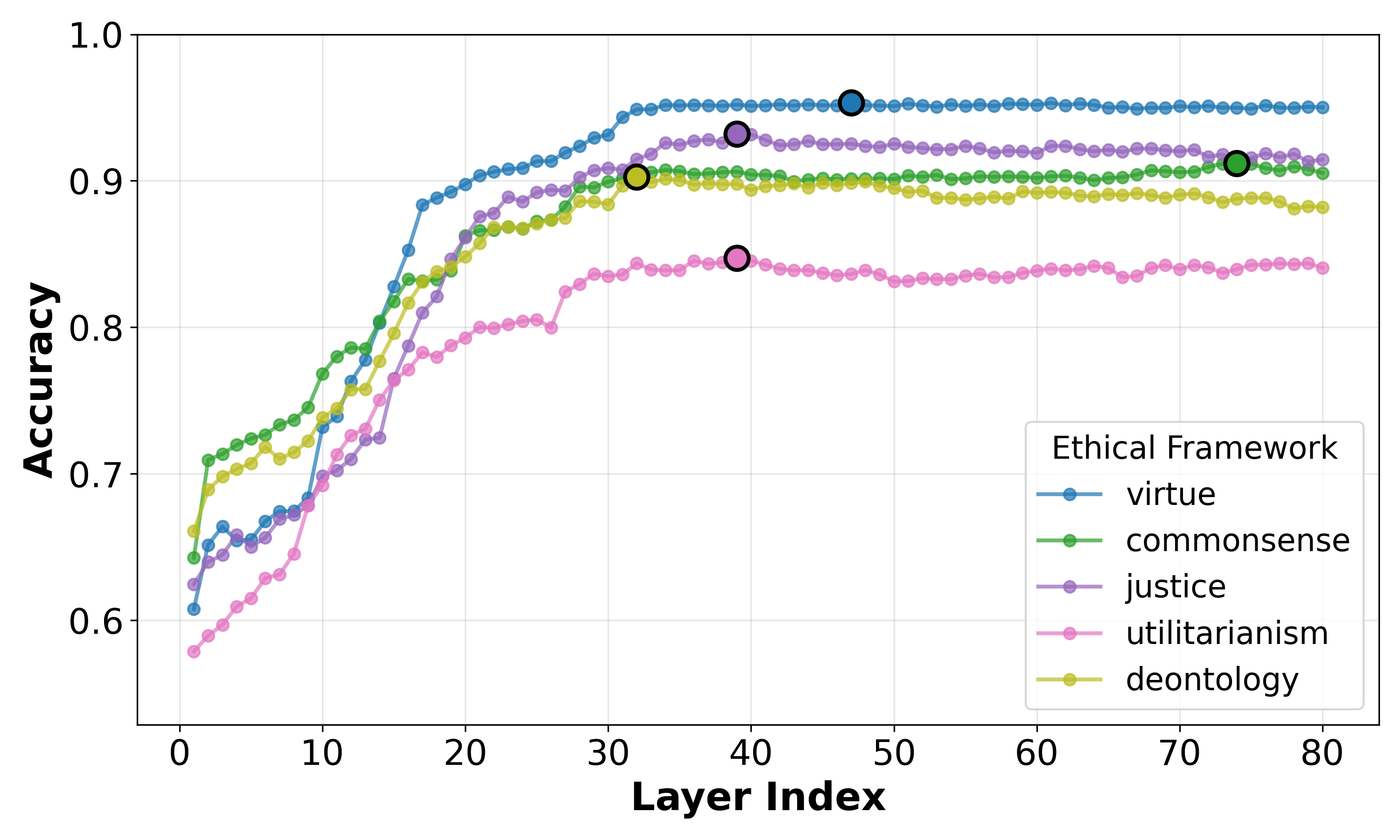} \caption{Within-framework accuracy emerges progressively, with optimal performance concentrated in the final third. Note the performance gap between virtue ethics (top) and utilitarianism (bottom).} \label{fig:layer_wise} \end{subfigure} \hfill \begin{subfigure}[t]{0.48\textwidth} \centering \raisebox{0.01\height}{\includegraphics[width=0.8\linewidth]{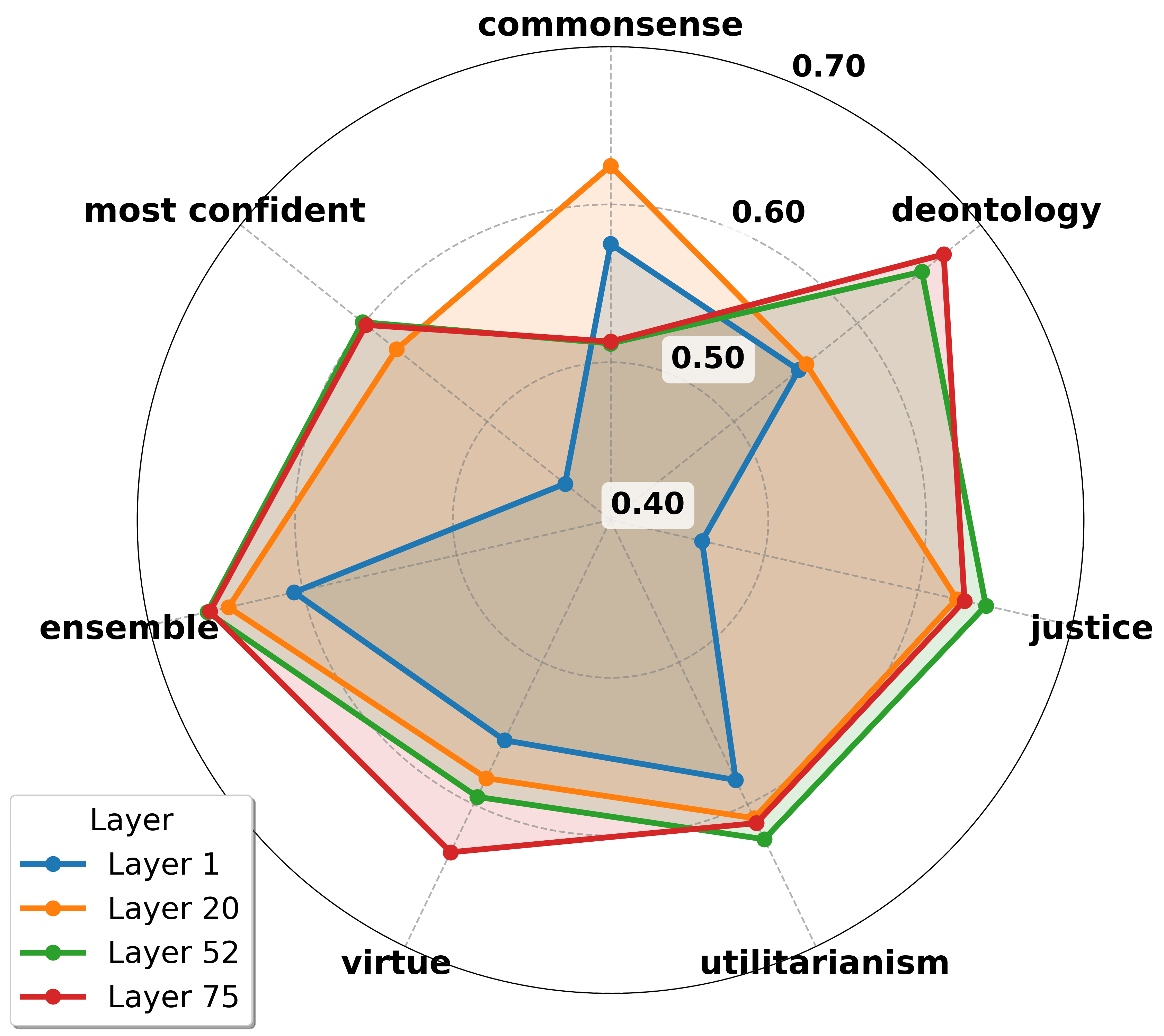}} \caption{Cross-framework generalization at representative layers. Irregular shapes visualize asymmetric transfer; quantitative claims rest on the accuracy values plotted, not visual geometry.} \label{fig:radar} \end{subfigure} \caption{Layer-wise emergence and cross-framework structure in Llama-3.3-70B-Instruct. (a) Framework-specific trajectories reveal a hierarchy of learnability. (b) Asymmetric transfer patterns confirm distributed, non-modular encoding.} \label{fig:probe_performance} \end{figure*}

\subsection{Probing Internal Conflict to Predict Behavioral Variability}

We hypothesize that when framework-specific representations strongly disagree, this tension manifests as behavioral inconsistency. We operationalize this through probe-based conflict detection using linear classifiers at 90\% model depth. Unlike the cross-framework analysis (65\% depth), here we use 90\% depth to capture the maximally specialized representations closest to generation.

We compute conflict scores as the product of probe disagreement and minimum confidence:
\begin{equation}
C = |p_d - p_u| \times \min(c_d, c_u)
\end{equation}
where $p_d, p_u$ are predicted probabilities and $c$ denotes confidence (distance from uncertainty, $2|p - 0.5|$).

This score captures scenarios where deontology and utilitarianism strongly and confidently disagree. If either probe is uncertain, the conflict score is downweighted, ensuring we flag genuine ethical dilemmas rather than general uncertainty.

To validate behavioral impact, we sample scenarios from high-conflict ($\ge$75th percentile) and low-conflict ($\leq$25th percentile) groups, generate 10 responses per prompt ($T=1.2$), and measure choice entropy. If representational conflict is functionally relevant, high-conflict inputs should produce significantly higher entropy.

\section{Results}

\begin{figure*}[!t]
\centering
\includegraphics[width=\linewidth]{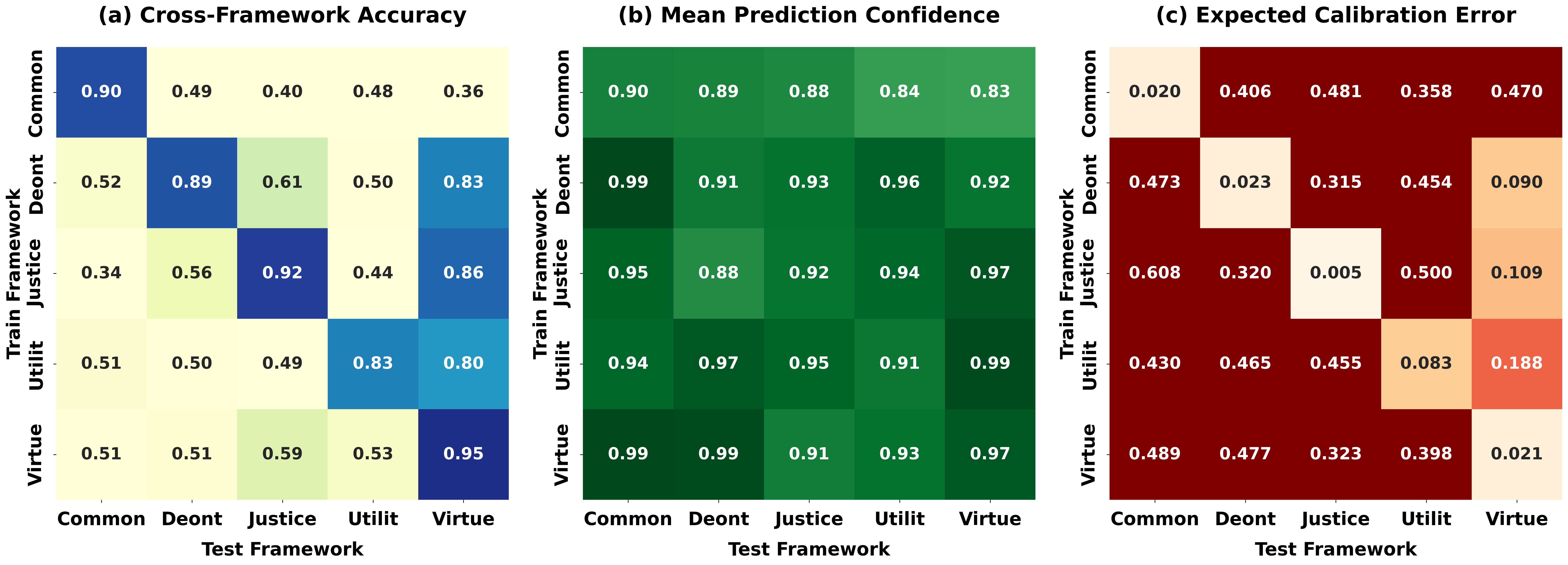}
\caption{Cross-framework probe analysis for Llama-3.3-70B-Instruct. (a) Accuracy matrices reveal asymmetric transfer. (b) Confidence heatmaps show pervasive overconfidence. (c) ECE confirms severe probe miscalibration on off-diagonal transfers, indicating that representations for different frameworks share structural features that trigger confident (but incorrect) predictions.}
\label{fig:cross_framework}
\end{figure*}

\subsection{Layer-wise Emergence and Specialization} \label{sec:layerwise_results}

Llama-3.3-70B reveals that ethical representations are not monolithic; they exhibit a clear hierarchy of learnability and depth-dependent specialization.

Virtue ethics representations achieve near-ceiling accuracy (95\%) while utilitarianism plateaus significantly lower (84\%), a disparity persisting from mid-layers onward (Figure~\ref{fig:layer_wise}). This performance gap suggests an architectural bias: the model more readily encodes static character-based patterns (virtue) than the complex, outcome-contingent evaluations required for utilitarianism. The trajectories also diverge temporally: Commonsense morality shows a steady, monotonic ascent, implying progressive integration of signals, whereas deontology peaks early (Layer 30) before stabilizing.

The ``plateau'' observed in mid-layers admits a structural interpretation. In smaller models (4B--7B), accuracy saturates by layer 20 (Appendix~\ref{app:layerwise}), marking a hard capacity constraint where ethical features are fully resolved relative to the model's limited expressivity. In contrast, frontier models like Llama-70B sustain gradual refinement through Layer 60. This extended maturation phase in larger models suggests that while early layers perform ``semantic extraction'' \cite{geva2022transformer}, the additional depth in 70B-scale architectures allows for nuance and conflict resolution that smaller models structurally lack.

The cross-framework radar charts (Figure~\ref{fig:radar}) visualize transfer accuracy patterns at representative layers. If models learned a unified acceptability dimension, transfer accuracy would form uniform shapes across axes. Instead, we observe asymmetric patterns—quantified formally in the transfer matrices (Figure~\ref{fig:cross_framework}a): justice and deontology exhibit substantial early-to-late gaps, showing that as representations become more specialized, they lose cross-framework generalization. The ``most confident'' selector consistently underperforms the ensemble average, a sign of systematic probe miscalibration as specialization increases.

These structural properties—hierarchical learnability, depth-dependent refinement, and asymmetric specialization—imply that alignment is not a uniform process. Interventions targeting ``easier'' frameworks like Virtue may not propagate to ``harder'' ones like Utilitarianism, and optimizing for specialization may inadvertently degrade the model's ability to generalize moral reasoning across domains.

\subsection{Cross-Framework Entanglement and Calibration}

Ethical representations are neither fully modular nor fully collapsed; they are entangled in ways that create specific reliability risks. Evaluating probes at 65\% depth (Layer 52 for Llama-3.3-70B), we isolate key structural properties.

The transfer matrices (Figure~\ref{fig:cross_framework}a) disprove the hypothesis that models learn a generic ``good/bad'' dimension. Deontology and utilitarianism encode abstract, generalizable features achieving moderate transfer success, while justice and commonsense appear highly specialized, failing catastrophically when transferred (e.g., Commonsense$\rightarrow$Justice approaches chance).

The sharpest finding concerns the dissociation between accuracy and confidence (Figure~\ref{fig:cross_framework}b). Probes maintain uniformly high confidence even during severe transfer failures. Because probe confidence reflects representational similarity to training examples, this indicates that representations for different frameworks share structural features sufficient to trigger confident (but incorrect) predictions. The model's representations do not signal ``this is an unusual input''—they lack the structure that would enable reliable out-of-distribution detection (Figure~\ref{fig:cross_framework}c).

This shared representational infrastructure complicates alignment: because frameworks share entangled features, optimization methods cannot target one framework in isolation. The persistence of this pattern across scales (4B--72B; Appendix~\ref{app:entanglement}) suggests it is an intrinsic transformer property rather than a model-specific artifact.

\subsection{Probe Conflict and Behavioral Inconsistency}
\label{sec:conflict}

\begin{figure}[!t]
  \centering
  \includegraphics[width=0.99\linewidth]{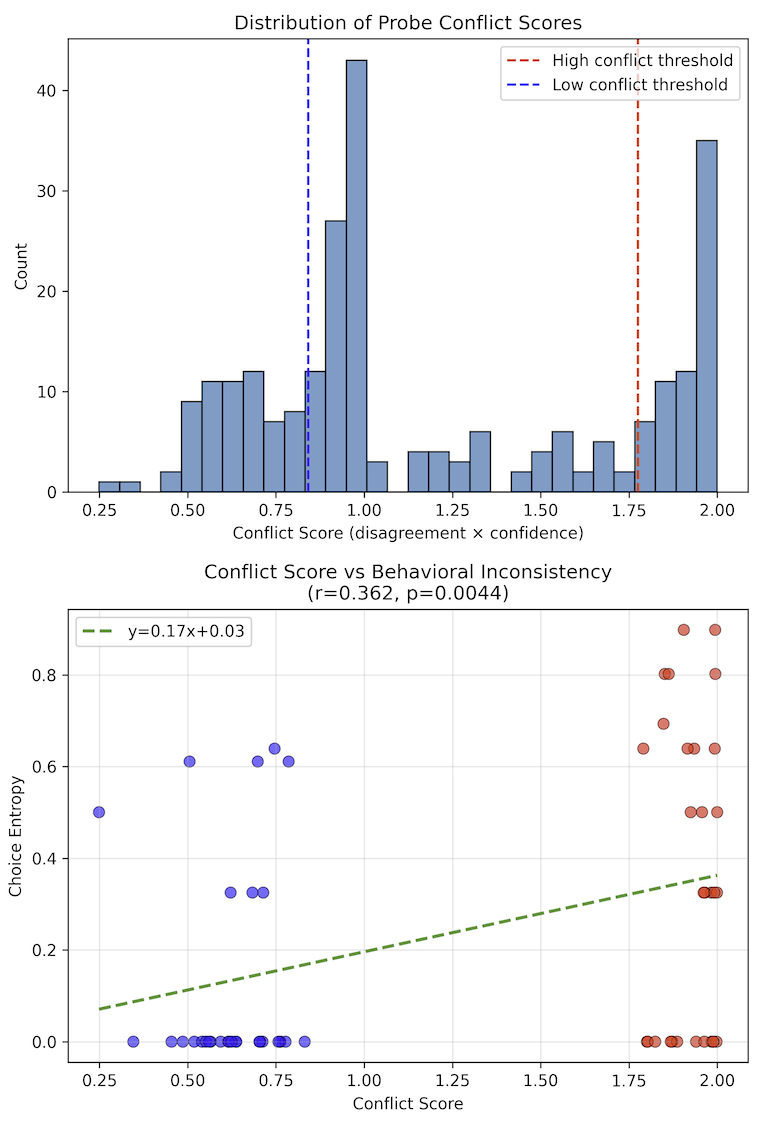}
\caption{Probe conflict and behavioral inconsistency in Mistral-7B-Instruct. (Top) The conflict distribution shows a populated high-conflict tail, indicating the model preserves distinct deontological and utilitarian representations that can effectively ``disagree.'' (Bottom) High-conflict scenarios (red) correlate with elevated choice entropy ($r = 0.362$, $p = 0.004$). The relationship may partly reflect shared sensitivity to scenario difficulty (see Limitations).}
\label{fig:conflict_analysis}
\end{figure}

We hypothesize that behavioral inconsistency may partly reflect unresolved tensions between internal ethical representations, though alternative explanations remain viable. To test this, we analyzed scenarios where deontological and utilitarian probes at 90\% depth made confident opposing predictions. We focus on Mistral-7B-Instruct as our primary case study due to its interpretable conflict distribution, though the findings replicate across all evaluated architectures.

The conflict distribution in Mistral-7B (Figure~\ref{fig:conflict_analysis}, top) reveals a noteworthy structural property: the model preserves distinct deontological and utilitarian subspaces rather than collapsing them into a single ``acceptability'' dimension. Unlike heavily aligned models that may suppress this differentiation (Appendix~\ref{appendix:llama_conflict}), Mistral maintains populated regions of both low and high conflict. This demonstrates that the model encodes competing ethical signals as distinct geometric vectors that can pull the decision process in opposing directions.

This internal differentiation has behavioral correlates. As shown in Figure~\ref{fig:conflict_analysis} (bottom), scenarios with high probe conflict exhibit significantly greater choice entropy than low-conflict ones ($r = 0.362$, $p = 0.004$). However, we note that this correlation has not been controlled for scenario difficulty---if more difficult scenarios independently produce both higher probe disagreement and higher behavioral entropy, the observed relationship may be partly or wholly confounded. Partial correlation analyses controlling for difficulty-related baselines (e.g., perplexity, logit margin) are needed to isolate the unique contribution of representational conflict. This correlation is consistent with but does not establish the hypothesis that when the model's internal representations strongly disagree, the model essentially flips a coin in its final output. High-conflict scenarios maintained high linguistic coherence, ruling out the alternative explanation that entropy stems from general confusion or linguistic failure; the model remains fluent in form but inconsistent in judgment.

\begin{table}[t]
\centering
\small
\setlength{\tabcolsep}{4pt}
\begin{tabular}{lcccc}
\toprule
\textbf{Model} & $r$ & \textbf{95\% CI} & $p$ & $d$ \\
\midrule
Mistral-7B & .362 & [.12, .56] & .004 & .58 \\
Qwen-2.5-7B & .290 & [.16, .41] & $<$.001 & .57 \\
Llama-3.3-70B & .317 & [.07, .53] & .014 & .54 \\
\bottomrule
\end{tabular}
\caption{Conflict-entropy relationship across model families. The correlation replicates across architectures with consistent medium magnitude ($d \approx 0.54$--$0.58$), surviving Bonferroni correction ($\alpha = 0.017$), though confound controls have not been applied.}
\label{tab:conflict_replication}
\end{table}

This conflict-entropy relationship replicates across architectures, not just within a single model. As detailed in Table~\ref{tab:conflict_replication}, the effect replicates across Qwen-7B ($n=200$) and Llama-70B ($n=62$) with consistent medium effect sizes (Cohen's $d \approx 0.54$--$0.58$). All correlations remain significant after Bonferroni correction. The relationship is slightly stronger in 7B models than in Llama-70B, suggesting that while heavy alignment training may mask the behavioral symptoms of conflict, the underlying representational tension persists.

While these correlations are suggestive, establishing practical utility for probe-based conflict detection requires demonstrating that the signal persists after controlling for simpler baselines such as scenario difficulty and output-level uncertainty. The conflict score's computational efficiency---requiring only a single forward pass---makes it a promising candidate for pre-screening, but its unique predictive contribution beyond difficulty-driven confounds remains to be validated.

\section{Discussion}

Our investigation establishes that ethical representations in LLMs are neither fully modular nor fully collapsed; they form a structured, entangled landscape with potential consequences for behavioral reliability.

\paragraph{Differentiated but Entangled Geometry.}
The asymmetric transfer patterns (Figure~\ref{fig:cross_framework}a) provide strong evidence against a simple ``good/bad'' encoding. Broad frameworks like deontology maintain partial generalization (0.67 accuracy to virtue), while narrow ones like commonsense fail catastrophically (0.51 to justice). However, this differentiation is imperfect: probes maintain high confidence even when transferred to inappropriate frameworks (Figure~\ref{fig:cross_framework}b), implying scenarios simultaneously activate multiple entangled patterns. This entanglement creates a persistent challenge for alignment: because frameworks share load-bearing features, optimization methods (RLHF, SFT, DPO) that modify late-layer representations to improve one framework risk distorting others. Strengthening justice may cannibalize features required by deontology, causing unpredictable degradation.

\paragraph{Conflict Correlates with Brittleness.}
Our conflict-entropy analysis is consistent with a possible mechanism for behavioral inconsistency, though confounding factors have not been ruled out. The correlation ($r \approx 0.36$) between probe disagreement and choice entropy suggests that ethical decisions may emerge from ``competition'' between internal signals---when deontological and utilitarian representations pull in opposite directions, the model becomes stochastic. However, scenario difficulty may independently drive both probe disagreement and behavioral entropy. This effect is stronger in smaller models than Llama-70B ($r=0.31$), suggesting that heavy alignment may mask conflicts without resolving them, though further controlled analyses are needed to confirm this interpretation.

\paragraph{Cross-Model Heterogeneity.}
These patterns manifest differently across architectures (detailed analyses in Appendices~\ref{app:layerwise}--\ref{app:entanglement}). Qwen exhibits scale-invariant virtue isolation—the 72B model replicates the 7B pattern almost identically, suggesting parameter scaling does not resolve structural entanglements. DeepSeek-R1 shows unique commonsense failures potentially reflecting cultural training differences from the Western-centric ETHICS benchmark. Llama-70B displays zero-conflict concentration that may reflect heavy alignment compression. While the specific mechanisms behind these differences require further investigation, the heterogeneity itself indicates that safety is not a monotonic function of scale—probe-based auditing must be calibrated to each architecture's structural fingerprint.

\paragraph{Potential Advantages and Methodological Caveats.}
Probe-based methods offer several potential advantages over behavioral evaluations, pending further validation. First, they may detect representational miscalibration---high probe confidence despite poor cross-domain accuracy (ECE $> 0.4$) suggests that representations share features across frameworks, creating risks where the model's internal structure provides no warning signal. Second, by detecting high-conflict internal states, probes could potentially flag inputs prone to inconsistency before deployment \cite{du2025investigating}. However, post-hoc validation (see Limitations) reveals that probes partially rely on surface lexical features and that probe-identified directions do not causally influence ethical accuracy via activation steering. These findings indicate that the ``advantages'' described here remain potential rather than established, and that probe-based methods require more rigorous validation before they can serve as reliable diagnostic tools.

\paragraph{Intervention Windows.}
Our layer-wise analysis (Figure~\ref{fig:probe_performance}a) maps territory for future interventions. Regardless of scale (4B--72B), every model exhibits peak cross-framework generalization at approximately 65\% depth, suggesting this transition from semantic encoding to specialized prediction is an intrinsic transformer property. Interventions at 60--70\% depth offer a strategic window where frameworks are differentiated but not yet hyper-specialized, though our preliminary steering experiments suggest that current probe-derived directions may not be suitable for this purpose (see Limitations).

\section{Conclusion}

We introduced a probing methodology examining whether conflicts between framework-specific probes correlate with behavioral inconsistency in LLMs. Three key findings emerge: (1) asymmetric transfer patterns rule out a collapsed single dimension, (2) severe miscalibration reveals entangled representations preventing reliable framework discrimination, and (3) representational conflicts correlate with behavioral inconsistency across architectures from 7B to 72B parameters.

Future work should validate the conflict-behavior link causally through activation interventions, conduct partial correlation analyses controlling for scenario difficulty, and develop training objectives that reduce representational entanglement. These findings suggest models maintain structured ethical representations whose relationship to behavioral consistency warrants further investigation with more rigorous controls.

\newpage

\section*{Limitations} 

\paragraph{Post-hoc validation and identified confounds.} After submission, we conducted three validation experiments that revealed important methodological limitations, which we report here for transparency.

\textit{Paraphrase sensitivity.} We generated semantically equivalent rephrasings of test scenarios and measured probe prediction consistency between original and paraphrased versions. Probe predictions showed low correlation across rephrasings, indicating that probes partially rely on surface lexical features specific to the ETHICS benchmark's template construction rather than underlying ethical content. This finding qualifies all claims about probes detecting ``ethical framework representations''---the detected patterns may reflect distributional signatures associated with each framework's dataset construction rather than genuine ethical encoding.

\textit{Activation steering dissociation.} We tested whether probe-identified directions could causally influence ethical judgments via activation steering. Steering along probe directions changed the model's position preference (A vs.\ B) but did not improve ethical accuracy. This dissociation suggests that the linear boundary identified by probes is not causally linked to moral reasoning, further supporting the interpretation that probes capture task-structural rather than ethical-content features.

\textit{Difficulty confound.} The conflict-entropy correlation reported in Section~\ref{sec:conflict} has not been controlled for scenario difficulty. Preliminary analysis suggests that more linguistically complex or ambiguous scenarios may independently produce both higher probe disagreement and higher behavioral entropy. Partial correlation analyses controlling for perplexity, logit margin, and output-level uncertainty are needed to determine whether representational conflict contributes unique predictive value beyond scenario difficulty. We report the uncorrected correlations and flag this as a priority for future validation.

Despite these limitations, several findings remain robust: the asymmetric cross-framework transfer patterns, layer-wise emergence trajectories, and entanglement structures are descriptive observations that do not depend on causal interpretation of probe signals. These structural properties of ethical representations merit further investigation with methods less susceptible to the confounds identified here.

\paragraph{Probe expressivity and causality.} Linear probes capture only linearly separable patterns, potentially missing non-linear ethical encodings. While we observe correlation between probe-detected conflicts and behavioral variability ($r \approx 0.3$--$0.36$), establishing causality requires activation interventions; our preliminary steering experiments (described above) suggest that probe directions do not causally govern ethical judgments.

\paragraph{Probing controls.} Our post-hoc validation experiments (described above) provide more informative controls than random-label baselines, revealing specific failure modes of linear probes on ethical content. Additionally, our experimental design provides implicit controls: asymmetric cross-framework transfer patterns are inconsistent with surface-cue memorization (if probes learned generic features, transfer would be uniform), and cross-architecture replication (7B--72B, four model families) rules out exploitation of model-specific artifacts. Our claims are diagnostic rather than ontological: we examine whether probe-accessible representational structure correlates with behavioral reliability, not that probes reveal ``genuine'' ethical understanding.

\paragraph{Dataset scope and generalization.} We analyze five frameworks from a single dataset family (ETHICS), which anchors findings to its annotation style, scenario framing, and Western cultural priors. We do not test on out-of-benchmark moral dilemmas, novel ethical scenarios, or cross-cultural datasets. This limits our ability to claim that the specific entanglement patterns (e.g., deontology-virtue overlap) generalize beyond ETHICS-style scenarios; they may partially reflect dataset-specific annotation artifacts rather than universal properties of ethical representation. However, several findings suggest broader relevance. First, the structural properties we identify—asymmetric transfer, high-confidence miscalibration, conflict-entropy coupling—are architectural patterns that should manifest regardless of specific scenario content. Second, our method demonstrated sensitivity to cultural variance where it existed: DeepSeek-R1-Distill-7B (Chinese-trained) exhibited unique commonsense failures that Western-trained models did not display, suggesting our approach can detect cultural-representational mismatches rather than being blind to them. Third, the ETHICS benchmark, while Western-centric in nomenclature, draws on philosophical distinctions (rules vs. consequences vs. character) that have analogs across ethical traditions. Future work should validate these patterns on diverse moral reasoning benchmarks and culturally varied scenarios to assess true generality.

\paragraph{Conflict detection and baseline comparisons.} Our analysis focuses on deontology-utilitarianism conflict at a single temperature ($T=1.2$) with 10 samples per prompt. Alternative framework pairs, multi-framework detection, or different generation parameters might reveal different relationships. Preliminary investigation suggests the conflict-entropy relationship may be confounded by scenario difficulty (see above). Systematic partial correlation analyses controlling for output-level uncertainty remain as future work. Even if simpler uncertainty measures achieve comparable predictive performance, probe-based conflict detection offers a distinct interpretability advantage: it identifies \textit{which} ethical frameworks are in tension, enabling targeted diagnosis that generic uncertainty metrics cannot provide.

\paragraph{Calibration interpretation.} Our ECE analysis measures probe calibration under distribution shift, not the model's internal epistemic uncertainty directly. High off-diagonal ECE indicates that probe confidence (reflecting representational similarity to training examples) remains high even when accuracy degrades—but this is a property of how probes decode representations, not necessarily the model's ``awareness'' of uncertainty. We interpret high probe confidence as evidence that representations for different frameworks share structural features (triggering confident predictions), rather than claiming direct access to the model's epistemic state. This interpretation is supported by the behavioral grounding: if probe confidence were merely an artifact of probe training, it should not predict downstream behavioral entropy. The fact that probe-identified conflicts do predict behavioral variability suggests the confidence signals reflect genuinely decision-relevant representational structure, though alternative interpretations remain possible.

\paragraph{Generalization.} While patterns persist across 4B--72B parameters and multiple architectures, findings may not extend to mixture-of-experts, retrieval-augmented systems, or constitutional AI models. We also characterize entanglement without tracing origins to training data, as pretraining recipes remain opaque.

\paragraph{Visualization and quantification.} Our UMAP projections and radar charts serve as qualitative illustrations; all quantitative claims rest on probe accuracy, transfer matrices, and statistical tests rather than visual patterns. Future work could employ formal separability measures (e.g., CKA, linear discriminability indices) to more rigorously quantify framework relationships beyond transfer accuracy. Such measures might reveal finer-grained structure that probe accuracy alone cannot capture.

\section*{Acknowledgments}
We used AI assistants for code generation and manuscript preparation. All experimental design, analysis, and scientific conclusions are the authors' original work.

\bibliography{custom}

\begin{thebibliography}{38}
\providecommand{\natexlab}[1]{#1}

\bibitem[{Azaria and Mitchell(2023)}]{azaria2023internal}
Amos Azaria and Tom Mitchell. 2023.
\newblock The internal state of an llm knows when it's lying.
\newblock \emph{EMNLP Findings}.

\bibitem[{Belinkov(2022)}]{belinkov2022probing}
Yonatan Belinkov. 2022.
\newblock Probing classifiers: Promises, shortcomings, and advances.
\newblock \emph{Computational Linguistics}, 48(1):207--219.

\bibitem[{Belinkov et~al.(2017)Belinkov, Shi, Glass, and
  Barzilay}]{belinkov2017neural}
Yonatan Belinkov, Xing Shi, James Glass, and Regina Barzilay. 2017.
\newblock What do neural machine translation models learn about morphology?
\newblock In \emph{Proceedings of the 55th Annual Meeting of the Association
  for Computational Linguistics}, pages 861--872.

\bibitem[{Burns et~al.(2023)Burns, Ye, Klein, and
  Steinhardt}]{burns2023discovering}
Collin Burns, Haotian Ye, Dan Klein, and Jacob Steinhardt. 2023.
\newblock Discovering latent knowledge in language models without supervision.
\newblock \emph{arXiv preprint arXiv:2212.03827}.

\bibitem[{Choi et~al.(2024)Choi, Kim, and Lee}]{choi2024moral}
Jongchan Choi, Mingyu Kim, and Sungju Lee. 2024.
\newblock Moral instruction fine-tuning for aligning {LMs} with multiple
  ethical principles.
\newblock In \emph{Proceedings of the 2024 IEEE International Conference on Big
  Data (BigData)}, pages 8647--8649. IEEE.

\bibitem[{Du(2025)}]{du2025investigating}
... Du. 2025.
\newblock Investigating value-reasoning reliability in small large language
  models.
\newblock In \emph{EMNLP}.

\bibitem[{Farquhar et~al.(2024)Farquhar, Kossen, Kuhn, and
  Gal}]{farquhar2024detecting}
Sebastian Farquhar, Jannik Kossen, Lorenz Kuhn, and Yarin Gal. 2024.
\newblock Detecting hallucinations in large language models using semantic
  entropy.
\newblock \emph{Nature}.

\bibitem[{Fedorenko et~al.(2024)Fedorenko, Piantadosi, and
  Gibson}]{fedorenko2024language}
Evelina Fedorenko, Steven~T Piantadosi, and Edward~AF Gibson. 2024.
\newblock Language is primarily a tool for communication rather than thought.
\newblock \emph{Nature}.

\bibitem[{Ganguli et~al.(2023)Ganguli, Askell, Schiefer, Liao
  et~al.}]{ganguli2023capacity}
Deep Ganguli, Amanda Askell, Nicholas Schiefer, Thomas Liao, and 1 others.
  2023.
\newblock The capacity for moral self-correction in large language models.
\newblock \emph{arXiv preprint arXiv:2302.07459}.

\bibitem[{{Gemma Team}(2025)}]{team2025gemma3}
{Gemma Team}. 2025.
\newblock Gemma 3.
\newblock \emph{arXiv preprint arXiv:2503.19786}.

\bibitem[{Geva et~al.(2022)Geva, Caciularu, Wang, and
  Goldberg}]{geva2022transformer}
Mor Geva, Avi Caciularu, Kevin Wang, and Yoav Goldberg. 2022.
\newblock Transformer feed-forward layers build predictions by promoting
  concepts in the vocabulary space.
\newblock In \emph{EMNLP}.

\bibitem[{Grattafiori et~al.(2024)Grattafiori, Dubey, Jauhri, Pandey, Kadian,
  Al-Dahle et~al.}]{grattafiori2024llama3}
Aaron Grattafiori, Abhimanyu Dubey, Abhinav Jauhri, Abhinav Pandey, Abhishek
  Kadian, Ahmad Al-Dahle, and 1 others. 2024.
\newblock The llama 3 herd of models.
\newblock \emph{arXiv preprint arXiv:2407.21783}.

\bibitem[{Guo et~al.(2025)Guo, Yang, Zhang et~al.}]{guo2025deepseek}
Daya Guo, Qihao Yang, Zhihao Zhang, and 1 others. 2025.
\newblock Deepseek-r1: Incentivizing reasoning capability in llms via
  reinforcement learning.
\newblock \emph{arXiv preprint arXiv:2501.00000}.

\bibitem[{Hendrycks et~al.(2021)Hendrycks, Burns, Basart, Critch, Li, Song, and
  Steinhardt}]{hendrycks2021aligning}
Dan Hendrycks, Collin Burns, Steven Basart, Andrew Critch, Jerry Li, Dawn Song,
  and Jacob Steinhardt. 2021.
\newblock Aligning ai with shared human values.
\newblock In \emph{International Conference on Learning Representations}.

\bibitem[{Hewitt and Liang(2019)}]{hewitt2019designing}
John Hewitt and Percy Liang. 2019.
\newblock Designing and interpreting probing classifiers.
\newblock In \emph{Proceedings of the 2019 Conference on Empirical Methods in
  Natural Language Processing}.

\bibitem[{Jiang et~al.(2023)Jiang, Sablayrolles, Mensch, Bamford
  et~al.}]{jiang2023mistral}
Albert~Q Jiang, Alexandre Sablayrolles, Arthur Mensch, Chris Bamford, and 1
  others. 2023.
\newblock Mistral 7b.
\newblock \emph{arXiv preprint arXiv:2310.06825}.

\bibitem[{Kadavath et~al.(2022)Kadavath, Conerly, Askell
  et~al.}]{kadavath2022language}
Saurav Kadavath, Tom Conerly, Amanda Askell, and 1 others. 2022.
\newblock Language models (mostly) know what they know.
\newblock \emph{arXiv:2207.05221}.

\bibitem[{Kosinski(2023)}]{kosinski2023theory}
Michal Kosinski. 2023.
\newblock Theory of mind may have spontaneously emerged in large language
  models.
\newblock \emph{arXiv preprint arXiv:2302.02083}.

\bibitem[{Kuhn et~al.(2023)Kuhn, Gal, and Farquhar}]{kuhn2023semantic}
Lorenz Kuhn, Yarin Gal, and Sebastian Farquhar. 2023.
\newblock Semantic uncertainty: Linguistic invariances for uncertainty
  estimation in natural language generation.
\newblock \emph{ICLR}.

\bibitem[{Li et~al.(2024)Li, Patel, Vi{\'e}gas, Pfister, and
  Wattenberg}]{li2024inference}
Kenneth Li, Oam Patel, Fernanda Vi{\'e}gas, Hanspeter Pfister, and Martin
  Wattenberg. 2024.
\newblock Inference-time intervention: Eliciting truthful answers from a
  language model.
\newblock \emph{NeurIPS}.

\bibitem[{Longpre et~al.(2021)Longpre, Periber, Chen, Dhingra, and
  Potts}]{longpre2021entity}
Shayne Longpre, Lu~Periber, Anthony Chen, Bhuwan Dhingra, and Christopher
  Potts. 2021.
\newblock Entity-based knowledge conflicts in question answering.
\newblock \emph{EMNLP}.

\bibitem[{Mahowald et~al.(2024)Mahowald, Ivanova, Blank, Kanwisher, Tenenbaum,
  and Fedorenko}]{mahowald2024dissociating}
Kyle Mahowald, Anna~A Ivanova, Idan~A Blank, Nancy Kanwisher, Joshua~B
  Tenenbaum, and Evelina Fedorenko. 2024.
\newblock Dissociating language and thought in large language models.
\newblock \emph{Trends in Cognitive Sciences}.

\bibitem[{Marraffini et~al.(2024)Marraffini, Cotton, Hsueh, Fridman, Wisznia,
  and Corro}]{marraffini2024greatest}
Giovanni Marraffini, Andr{\'e}s Cotton, Noe Hsueh, Axel Fridman, Juan Wisznia,
  and Luciano Corro. 2024.
\newblock The greatest good benchmark: Measuring llms’ alignment with
  utilitarian moral dilemmas.
\newblock In \emph{Proceedings of the 2024 Conference on Empirical Methods in
  Natural Language Processing}, pages 21950--21959.

\bibitem[{Neeman et~al.(2023)Neeman, Aharoni, Honovich, Choshen, Szpektor, and
  Abend}]{neeman2023disentqa}
Ella Neeman, Roee Aharoni, Or~Honovich, Leshem Choshen, Idan Szpektor, and Omri
  Abend. 2023.
\newblock Disentqa: Disentangling parametric and contextual knowledge with
  counterfactual question answering.
\newblock \emph{ACL}.

\bibitem[{Neuman and Shah(2025)}]{neuman2025analyzing}
W.~R. Neuman and Manan Shah. 2025.
\newblock Analyzing the ethical logic of six large language models.
\newblock \emph{arXiv}.

\bibitem[{Park et~al.(2024)Park, Goldstein, O'Gara, Chen, and
  Hendrycks}]{park2024ai}
Peter~S Park, Simon Goldstein, Aidan O'Gara, Michael Chen, and Dan Hendrycks.
  2024.
\newblock Ai deception: A survey of examples, risks, and potential solutions.
\newblock \emph{Patterns}, 5(5).

\bibitem[{Rimsky et~al.(2023)Rimsky, Gabrieli, Schulz, Tong, Hubinger, and
  Turner}]{rimsky2023steering}
Nina Rimsky, Nick Gabrieli, Julian Schulz, Meg Tong, Evan Hubinger, and
  Alexander~Matt Turner. 2023.
\newblock Steering llama-2 with contrastive activation additions.
\newblock \emph{arXiv preprint arXiv:2312.06681}.

\bibitem[{Scherrer et~al.(2024)Scherrer, Shi, Feder, and
  Blei}]{scherrer2024evaluating}
Nino Scherrer, Claudia Shi, Amir Feder, and David Blei. 2024.
\newblock Evaluating the moral beliefs encoded in llms.
\newblock In \emph{Proceedings of the 61st Annual Meeting of the Association
  for Computational Linguistics}, pages 3521--3539.

\bibitem[{Simmons and Hare(2022)}]{simmons2022moral}
Gabriel Simmons and Christopher Hare. 2022.
\newblock Moral mimicry: Large language models produce moral rationalizations
  tailored to political identity.
\newblock \emph{arXiv preprint arXiv:2209.12106}.

\bibitem[{Stoehr et~al.(2024)Stoehr, Cheng, Wang, Preotiuc-Pietro, and
  Bhowmik}]{stoehr2024unsupervised}
Niklas Stoehr, Pengxiang Cheng, Jing Wang, Daniel Preotiuc-Pietro, and Rajarshi
  Bhowmik. 2024.
\newblock Unsupervised contrast-consistent ranking with language models.
\newblock In \emph{Proceedings of the 18th Conference of the European Chapter
  of the Association for Computational Linguistics}, pages 483--495.

\bibitem[{Tanmay et~al.(2023)Tanmay, Khandelwal, Agarwal, and
  Choudhury}]{tanmay2023probing}
Kumar Tanmay, Aditi Khandelwal, Utkarsh Agarwal, and Monojit Choudhury. 2023.
\newblock Probing the moral development of large language models through
  defining issues test.
\newblock \emph{arXiv preprint arXiv:2309.13356}.

\bibitem[{Tenney et~al.(2019)Tenney, Das, and Pavlick}]{tenney2019bert}
Ian Tenney, Dipanjan Das, and Ellie Pavlick. 2019.
\newblock Bert rediscovers the classical {NLP} pipeline.
\newblock In \emph{Proceedings of the 57th Annual Meeting of the Association
  for Computational Linguistics}, pages 4593--4601.

\bibitem[{Tlaie(2024)}]{tlaie2024exploring}
Alejandro Tlaie. 2024.
\newblock Exploring and steering the moral compass of large language models.
\newblock \emph{arXiv}.

\bibitem[{Wang et~al.(2023)Wang, Wei, Schuurmans, Le, Chi, Narang, Chowdhery,
  and Zhou}]{wang2023selfconsistency}
Xuezhi Wang, Jason Wei, Dale Schuurmans, Quoc Le, Ed~Chi, Sharan Narang,
  Aakanksha Chowdhery, and Denny Zhou. 2023.
\newblock Self-consistency improves chain of thought reasoning in language
  models.
\newblock \emph{ICLR}.

\bibitem[{Yang et~al.(2024)Yang, Yang, Zhang, Hui, Zheng, Yu
  et~al.}]{yang2024qwen2}
An~Yang, Baosong Yang, Beichen Zhang, Binyuan Hui, Bo~Zheng, Bowen Yu, and 1
  others. 2024.
\newblock Qwen2.5 technical report.
\newblock \emph{arXiv preprint arXiv:2412.15115}.

\bibitem[{Yuan and Singh(2024)}]{yuan2024right}
Jiaqing Yuan and Munindar~P. Singh. 2024.
\newblock Right vs. right: Can llms make tough choices?
\newblock \emph{arXiv}.

\bibitem[{Ziems et~al.(2022)Ziems, Yu, Wang, Halevy, and Yang}]{ziems2022moral}
Caleb Ziems, Jane Yu, Yi-Chia Wang, Alon Halevy, and Diyi Yang. 2022.
\newblock The moral integrity corpus: A benchmark for ethical dialogue systems.
\newblock In \emph{Proceedings of the 60th Annual Meeting of the Association
  for Computational Linguistics}.

\bibitem[{Zou et~al.(2023)Zou, Phan, Chen, Campbell, Guo
  et~al.}]{zou2023representation}
Andy Zou, Long Phan, Sarah Chen, James Campbell, Phillip Guo, and 1 others.
  2023.
\newblock Representation engineering: A top-down approach to ai transparency.
\newblock \emph{arXiv preprint arXiv:2310.01405}.

\end{thebibliography}

\appendix

\section{Prompt Engineering}
\label{app:prompts}

\paragraph{Preprocessing.} The original ETHICS utilitarianism dataset contains pleasant/unpleasant scenario pairs. To prevent positional bias, we randomly assign scenarios to positions with equal probability: either A$=$pleasant and B$=$unpleasant (label$=$1), or A$=$unpleasant and B$=$pleasant (label$=$0). This randomization ensures that high probe accuracy reflects the model's ability to distinguish utilitarian preferences from scenario content rather than learning position-based shortcuts—if probes achieved above-chance accuracy by position memorization, random assignment would reduce accuracy to 50\%. We verified that train/test splits maintain balanced position distributions. Virtue ethics scenarios contain \texttt{[SEP]} tokens separating behaviors from traits, which we parse into structured components. All other frameworks require only standardized prompt formatting.

\paragraph{Prompt Templates.} Our prompts do not explicitly name ethical frameworks (e.g., ``Based on utilitarian principles, which scenario is preferable?''). This design is central to our research question: whether models have internalized distinct framework-associated patterns during pretraining and instruction tuning. Explicitly naming frameworks would invalidate our methodology in three ways. First, it shifts measurement from implicit representations to instruction-following ability—we would test whether models can apply labeled ethical lenses rather than whether they encode these frameworks internally. Second, it confounds probing analysis: if prompts announce ``use deontology,'' then cross-framework probe transfer no longer measures representational entanglement but merely whether models ignore instructions. Third, it introduces demand characteristics where models pattern-match framework keywords (e.g., ``utilitarian'' $\rightarrow$ ``maximize welfare'') rather than engage genuine ethical processing, causing probes to detect shallow heuristics instead of deep representational structure.

We use standardized templates across all scenarios within each framework to isolate representational geometry from prompt engineering effects. All prompts follow the format \texttt{[Context] $\rightarrow$ [Question] $\rightarrow$ Answer (A) or (B)}, but vary in their contextual focus and question framing. Utilitarian prompts present comparative scenarios and ask ``Which is preferable?'' to emphasize consequential evaluation. Deontological prompts provide excuses and scenarios, asking ``Should this scenario occur?'' to focus on rule permissibility. Commonsense prompts query ``Is this morally acceptable?'' to elicit social intuitions. Virtue prompts pair behaviors with traits, asking ``Does this behavior reflect this trait?'' to assess character-based reasoning. Justice prompts ask ``Is this scenario just?'' to evaluate fairness.

\section{Probing Implementation}
\label{app:probing}

\paragraph{Feature Extraction.} We extract hidden states from all transformer layers using the final non-padding token position. Inputs are tokenized with truncation at 512 tokens and \texttt{padding='longest'} to minimize computational overhead while preserving genuine token representations. This length limit captures 99.5\% of scenarios completely, with over-length cases concentrated in the commonsense framework. Features are verified to contain no NaN values before probe training.

\paragraph{Probe Architecture.} We implement logistic regression probes with L2 regularization ($C=0.01$), balanced class weighting, and LBFGS optimization (max 2000 iterations). Features are standardized to zero mean and unit variance using sklearn's \texttt{StandardScaler} fitted on training data.

\paragraph{Layer Selection.} We train probes across all transformer layers. For cross-framework experiments requiring a unified comparison depth, we select layer $\lfloor 0.65 \times n_{\text{layers}} \rfloor$. This selection is motivated by two empirical observations: (1) different frameworks achieve peak accuracy at different layers (e.g., virtue ethics peaks earlier than utilitarianism), so 65\% depth provides a unified point where all frameworks have achieved sufficient representational maturity; (2) cross-framework generalization consistently peaks at approximately 65\% depth across all evaluated architectures (4B--72B) before degrading into narrow specialization in final layers. This depth thus represents the optimal trade-off between representational maturity and preserved cross-domain generalizability, enabling valid comparison across framework pairs.

\section{Conflict Detection Implementation}
\label{app:conflict}

\paragraph{Probe Configuration.} We train deontology and utilitarianism probes on their respective training sets using the same architecture as cross-framework analysis: logistic regression with $C=0.01$, balanced class weighting, standardized features, evaluated at 90\% model depth (Layer 29 of 32 for Mistral-7B-Instruct; Layer 72 of 80 for Llama-3.3-70B-Instruct).

\paragraph{Confidence Calculation.} For each probe's predicted probability $p \in [0,1]$ for the positive class, we compute confidence as $c = 2|p - 0.5|$, measuring distance from maximum uncertainty (0.5) and normalizing to [0,1]. This ensures confidence reflects certainty regardless of which class is predicted.

\paragraph{Conflict Score.} We define conflict as:
\begin{equation}
C = |p_d - p_u| \times \min(c_d, c_u)
\end{equation}
where $c_d, c_u$ are deontology and utilitarianism probe confidences, and $p_d, p_u$ are their predicted probabilities. This formulation requires both strong disagreement between probes ($|p_d - p_u|$ approaching 1) and high confidence from both frameworks. The minimum operator ensures that conflict is downweighted when either probe is uncertain, isolating scenarios where both frameworks make confident opposing predictions.

\paragraph{Scenario Selection.} From the full test set, we select scenarios at the 75th percentile (high conflict) and 25th percentile (low conflict) of the conflict score distribution. For Mistral-7B and Llama-70B, we sample 100 scenarios from each group for response generation, yielding 200 scenarios total per model (after filtering for valid binary choice extraction, $n = 62$ per model for correlation analysis). For Qwen-2.5-7B, higher valid response rates yielded $n = 200$ analyzable scenarios.

\paragraph{Response Generation.} For each selected scenario, we generate 10 responses using temperature sampling ($T=1.2$, $\texttt{max\_new\_tokens=512}$, $\texttt{do\_sample=True}$). Prompts are formatted using model-specific instruction templates. Total generations: 2,000 per model (200 scenarios $\times$ 10 generations).

\paragraph{Choice Extraction.} We extract binary choices (A vs B) from generated responses using hierarchical pattern matching with priority ordering: (1) explicit choice markers ``(A)'' or ``(B)'', (2) answer prefixes matching ``Answer: [A/B]'', (3) responses beginning with ``A'' or ``B'', (4) first occurrence of A or B within the initial 100 characters, (5) phrases like ``Option A'' or ``Choose B''. Responses matching none of these patterns are labeled ``OTHER'', capturing refusals, explanations without clear choices, or malformed outputs. In our analysis, OTHER responses constituted less than 5\% of generations, indicating high compliance with binary choice prompts.

\paragraph{Behavioral Consistency Measurement.} We compute choice entropy from extracted binary decisions to measure behavioral variability. Shannon entropy is calculated over the choice distribution for each scenario: $H = -\sum_i p_i \log_2 p_i$ where $p_i$ is the proportion of each choice category across the 10 generations. Higher entropy indicates greater inconsistency, with maximum entropy (1.0 for binary choices) occurring when responses are evenly split between A and B.

\section{Layer-wise Emergence Across Model Scales}
\label{app:layerwise}

Progressive emergence of framework-specific representations generalizes across architectures and parameter counts, while scale primarily modulates refinement dynamics. Smaller models (4B--7B) plateau by layer~20 with peak accuracies around 0.83--0.93, whereas frontier-scale architectures (70B+) sustain improvement through layer~60, approaching near-ceiling performance around 0.95. The key distinction across scales lies not in representational structure but in depth-dependent refinement capacity, reflecting how additional layers extend and stabilize ethical subspaces without altering their intrinsic geometry.

\subsection{Small-Scale Model (4B Parameters)}

\begin{figure}[H]
  \centering
  \includegraphics[width=0.99\linewidth]{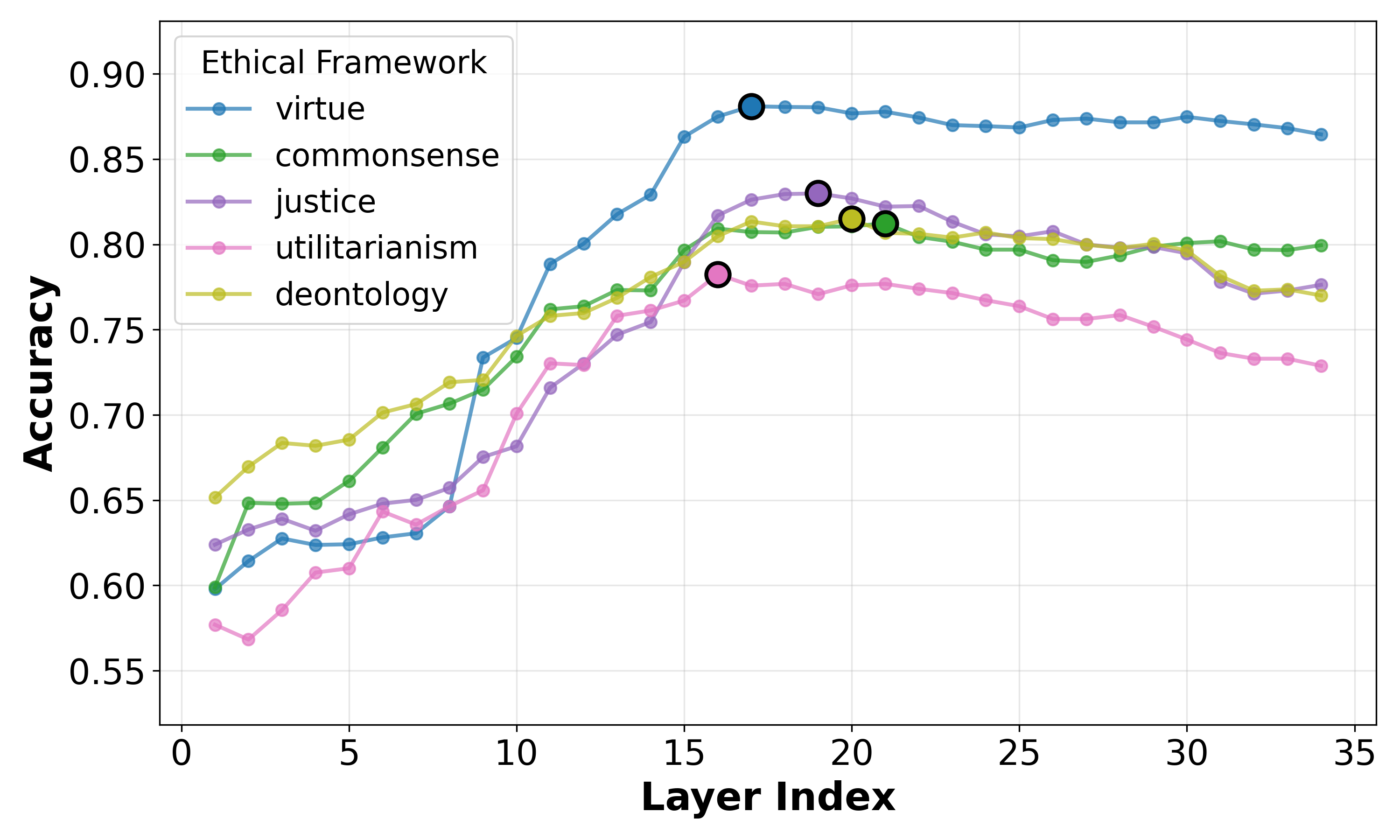}
  \caption{Layer-wise probe accuracy for Gemma-3-4B-Base across 34 layers. Framework-specific trajectories emerge despite limited capacity, with virtue ethics achieving early plateau while utilitarianism shows delayed convergence. The base model exhibits lower absolute performance but preserves relative framework ordering.}
  \label{fig:gemma_layers}
\end{figure}

Figure~\ref{fig:gemma_layers} presents layer-wise probe accuracy for Gemma-3-4B-Base. Virtue ethics peaks around layer~17, while deontology, justice, and commonsense frameworks converge more gradually, reaching maxima near layers~20--22. Utilitarianism remains consistently lower, indicating weaker separability. All trajectories plateau by layer~22, with minor decline thereafter, reflecting early saturation under limited capacity. Despite the absence of instruction tuning, accuracies across frameworks remain within 0.8--0.9, confirming that ethical distinctions emerge from pretraining alone.

\subsection{Medium-Scale Models (7B Parameters)}

\begin{figure}[!t]
  \centering
  \begin{subfigure}[t]{0.99\linewidth}
    \centering
    \includegraphics[width=\linewidth]{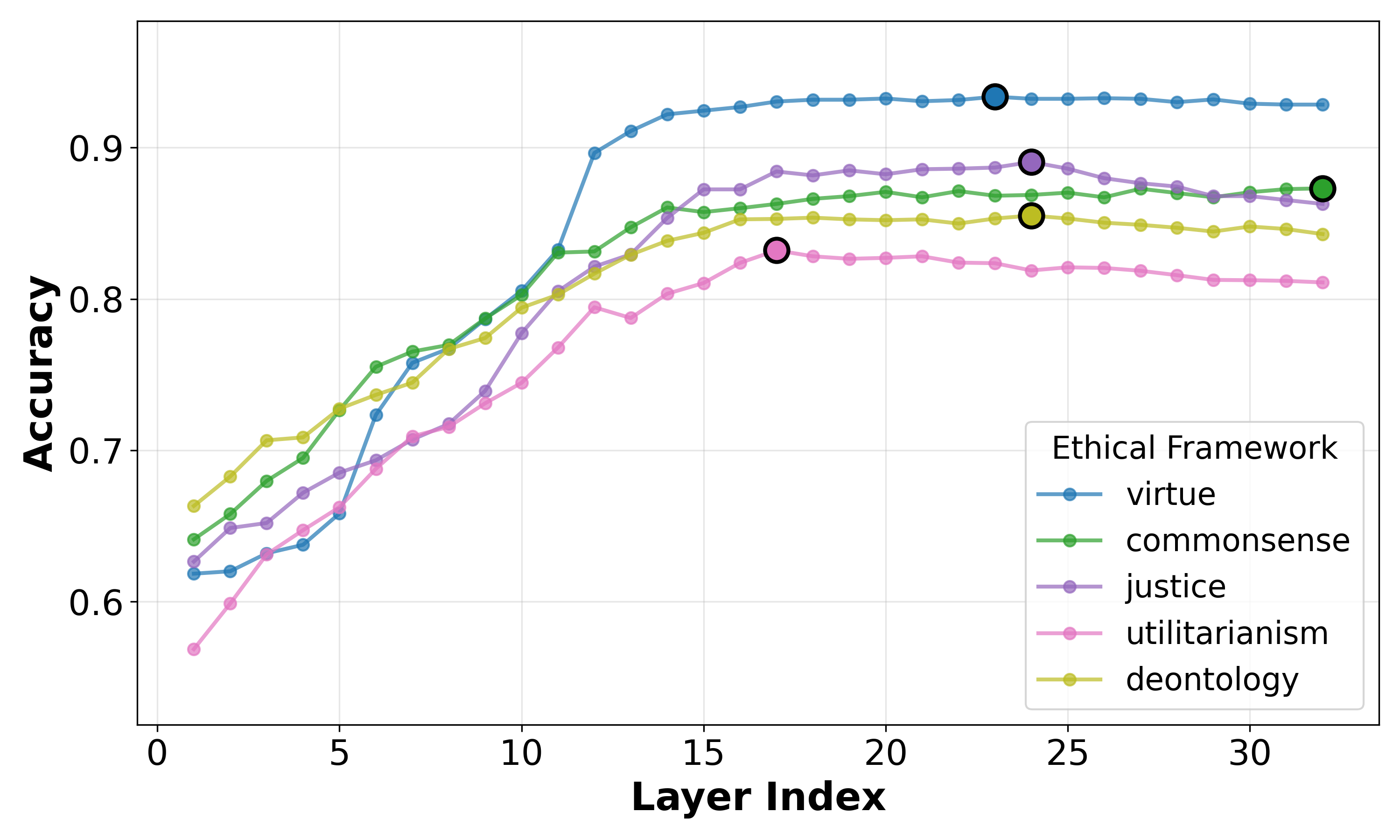}
    \caption{Mistral-7B-Instruct (32 layers)}
    \label{fig:mistral_layers}
  \end{subfigure}
  
  \vspace{0.3cm}
  
  \begin{subfigure}[t]{0.99\linewidth}
    \centering
    \includegraphics[width=\linewidth]{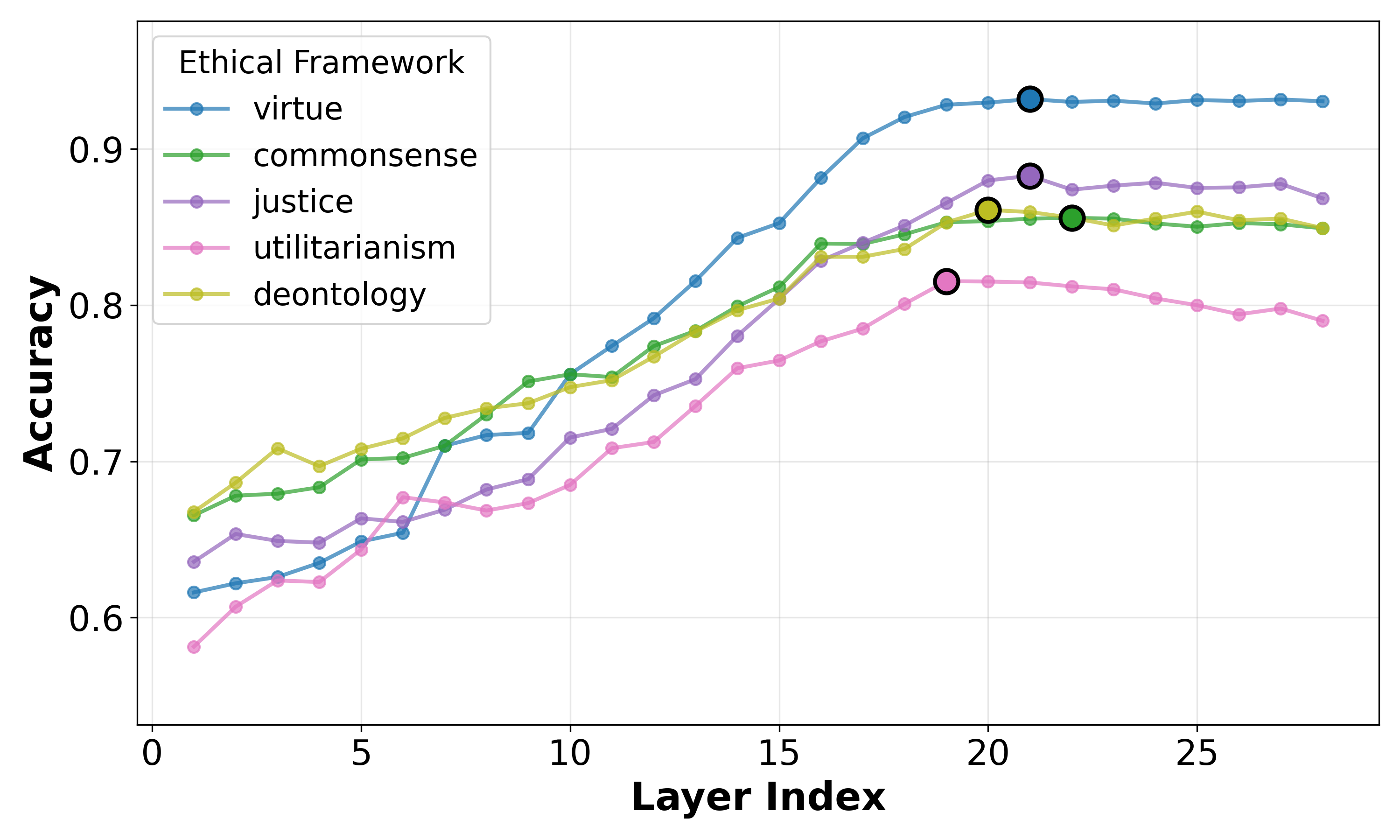}
    \caption{Qwen-2.5-7B-Instruct (28 layers)}
    \label{fig:qwen7b_layers}
  \end{subfigure}
  
  \vspace{0.3cm}
  
  \begin{subfigure}[t]{0.99\linewidth}
    \centering
    \includegraphics[width=\linewidth]{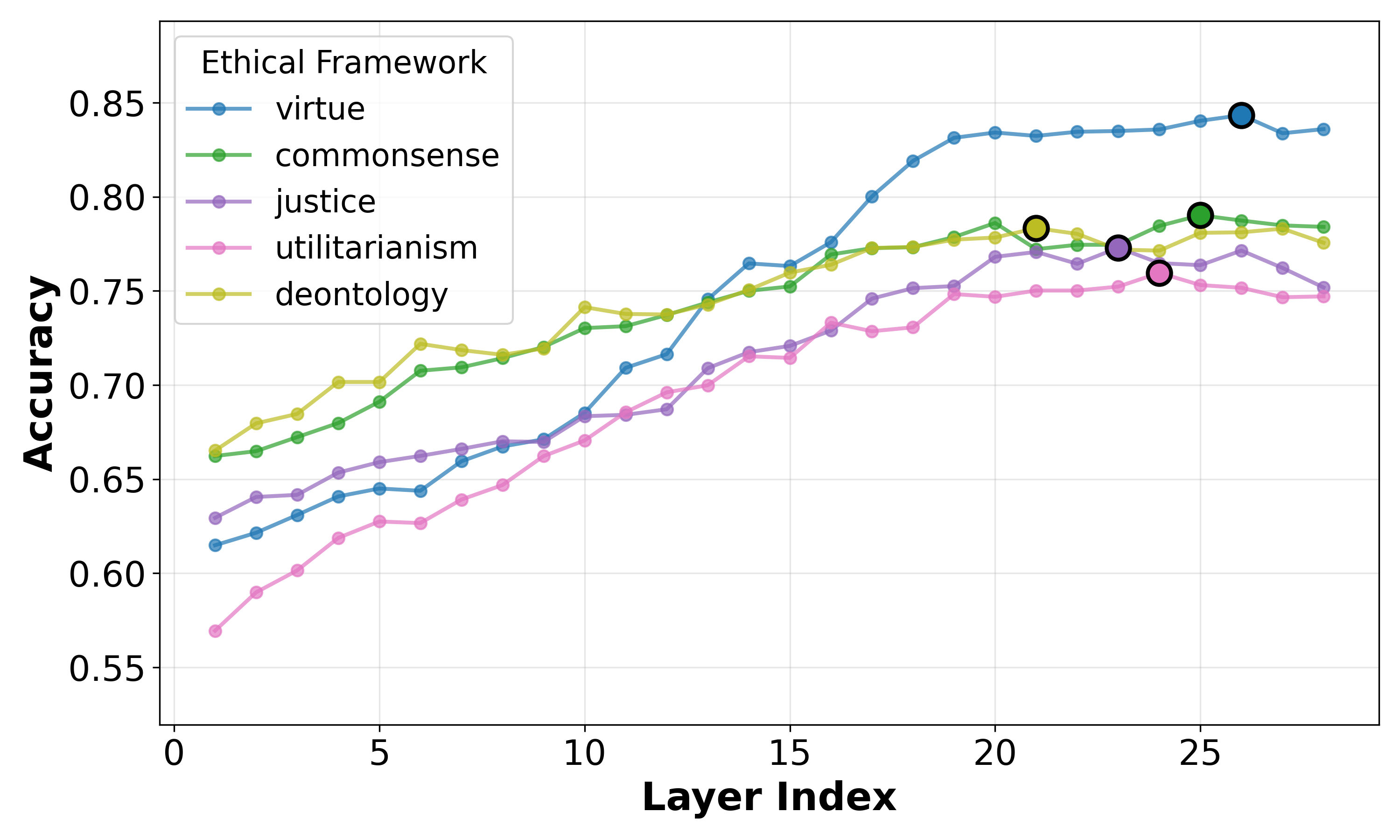}
    \caption{DeepSeek-R1-Distill-7B (28 layers)}
    \label{fig:deepseek_layers}
  \end{subfigure}
  
  \caption{Layer-wise probe accuracy across three 7B-parameter models with different training approaches: standard instruction tuning (Mistral, Qwen) versus reasoning-specialized distillation (DeepSeek-R1). All exhibit progressive emergence with framework-specific trajectories, though absolute performance varies with training methodology.}
  \label{fig:7b_comparison}
\end{figure}

Figure~\ref{fig:7b_comparison} compares layer-wise probe accuracy across three 7B-parameter models with distinct training paradigms. Across all models, framework representations emerge progressively through early layers (0--15) and stabilize thereafter. Compared with smaller 4B models, the 7B architectures sustain refinement deeper into the network, maintaining upward trends until layers~18--22 before plateauing. Instruction-tuned models (Mistral, Qwen) achieve higher peak accuracy than reasoning-distilled ones, implying that supervised alignment enhances discriminative sharpness without altering representational topology. The persistence of hierarchical ordering—virtue preceding justice, commonsense, and deontology—indicates that ethical differentiation arises primarily from architectural priors and pretraining exposure rather than explicit alignment signals.

\subsection{Large-Scale Models (70B+ Parameters)}

\begin{figure}[!t]
  \centering
  \begin{subfigure}[t]{0.99\linewidth}
    \centering
    \includegraphics[width=\linewidth]{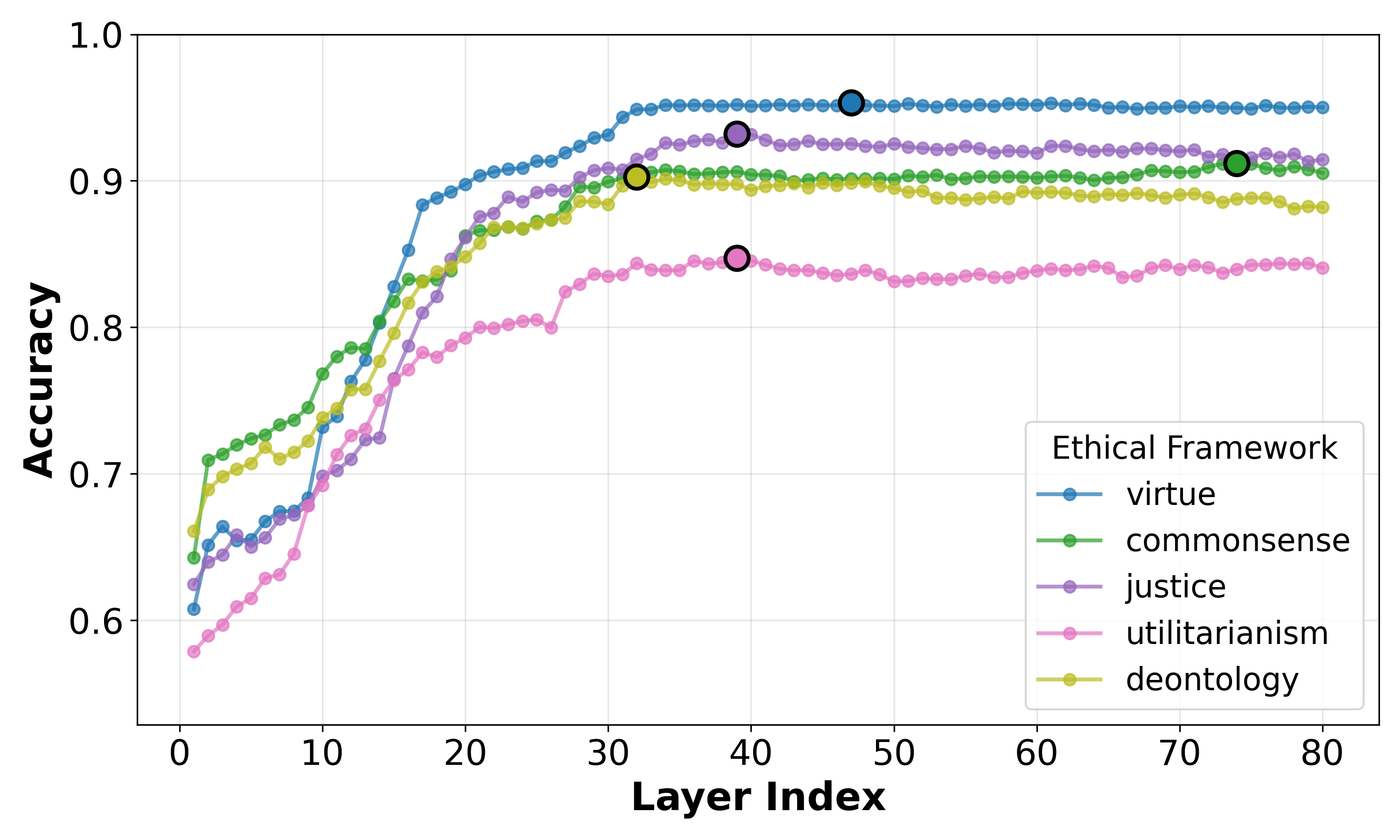}
    \caption{Llama-3.3-70B-Instruct (80 layers)}
    \label{fig:llama70b_layers}
  \end{subfigure}
  
  \vspace{0.3cm}
  
  \begin{subfigure}[t]{0.99\linewidth}
    \centering
    \includegraphics[width=\linewidth]{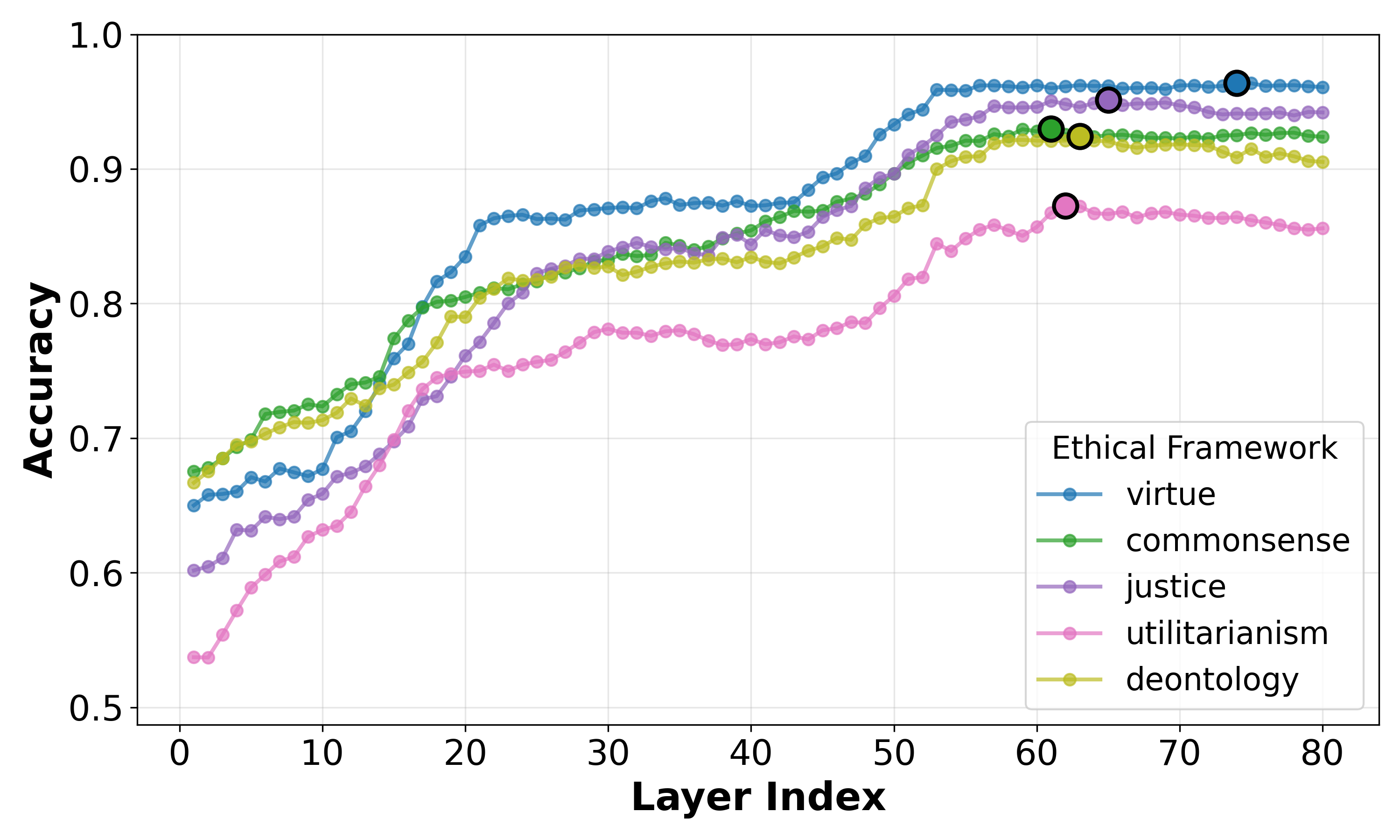}
    \caption{Qwen-2.5-72B-Instruct (80 layers)}
    \label{fig:qwen72b_layers}
  \end{subfigure}
  
  \caption{Layer-wise probe accuracy for frontier-scale models. Extended depth enables continued late-layer refinement, with some frameworks showing sustained improvement beyond the 50th layer. Higher absolute performance across all frameworks suggests increased capacity supports more nuanced ethical discrimination.}
  \label{fig:70b_comparison}
\end{figure}

Figure~\ref{fig:70b_comparison} illustrates how expanded depth alters representational dynamics in frontier-scale architectures. Unlike smaller models that saturate by layer~20, both 70B-scale systems sustain refinement through layer~60, revealing deeper stratification of ethical subspaces. Qwen-2.5-72B-Instruct displays a distinct two-phase trajectory—early consolidation by layer~30 followed by secondary gains after layer~50—suggesting layered specialization rather than uniform progression. The persistence of framework-specific ordering across scales—virtue leading, utilitarianism lagging—indicates that computational depth modulates precision rather than principle.

\section{Cross-Framework Generalization Across Model Scales}
\label{app:entanglement}

The non-monotonic generalization pattern replicates across all scales with notable consistency: cross-domain transfer peaks at middle-to-late depths (65\%) then degrades as framework-specific specialization dominates. Scale modulates both the sharpness of this trade-off and the absolute generalization ceiling, revealing how capacity constraints shape ethical representation structure.

\subsection{Small-Scale Models (4B Parameters)}

\begin{figure}[!htbp]
  \centering
  \includegraphics[width=0.85\linewidth]{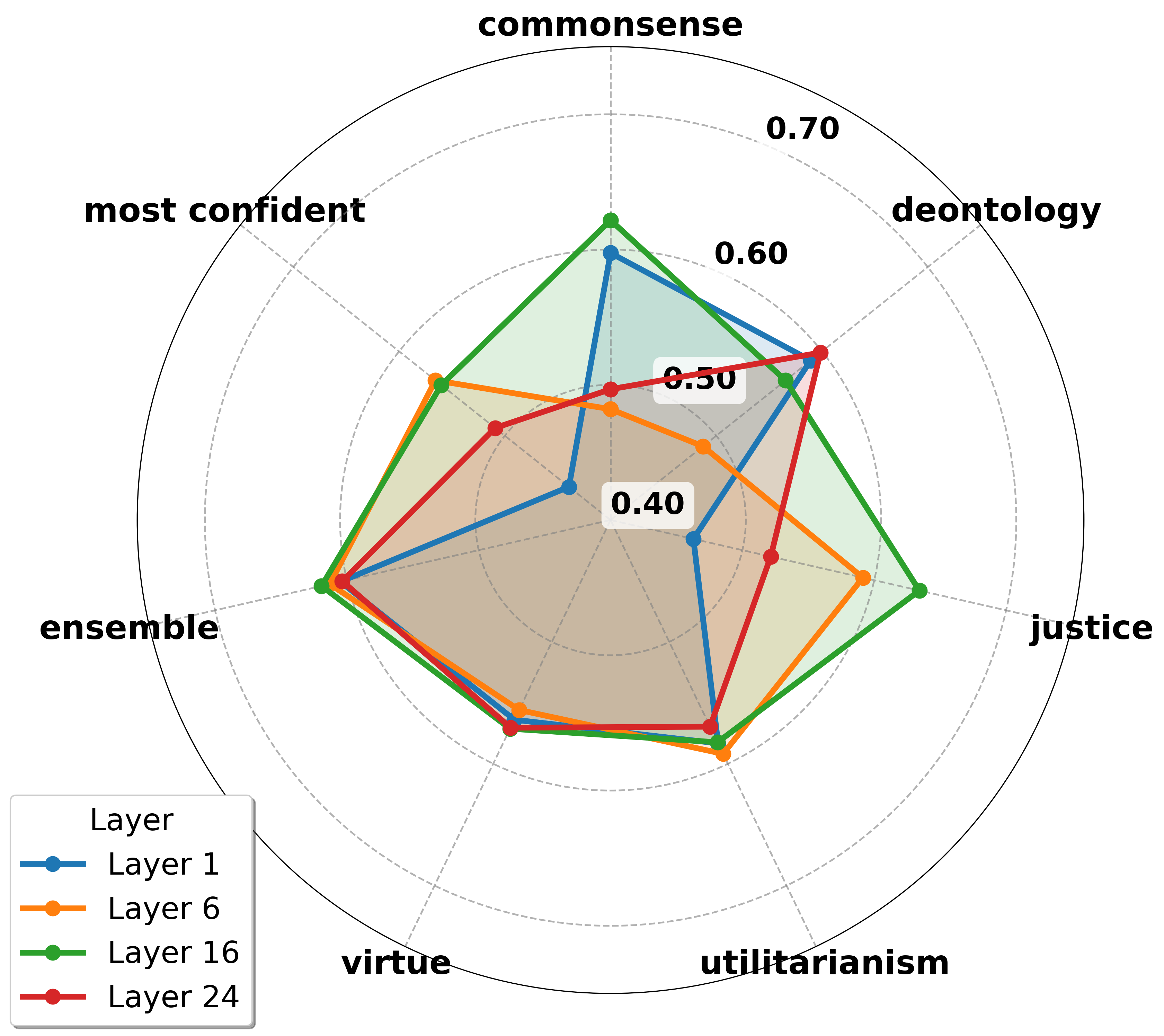}
  \caption{Cross-domain generalization for Gemma-3-4B-Base at four network depths (1\%, 25\%, 65\%, 94\%). Peak generalization occurs at 65\% depth, with sharp degradation by 94\% depth.}
  \label{fig:gemma_spider}
\end{figure}

\begin{figure}[!t]
  \centering
  \begin{subfigure}[t]{0.85\linewidth}
    \centering
    \includegraphics[width=\linewidth]{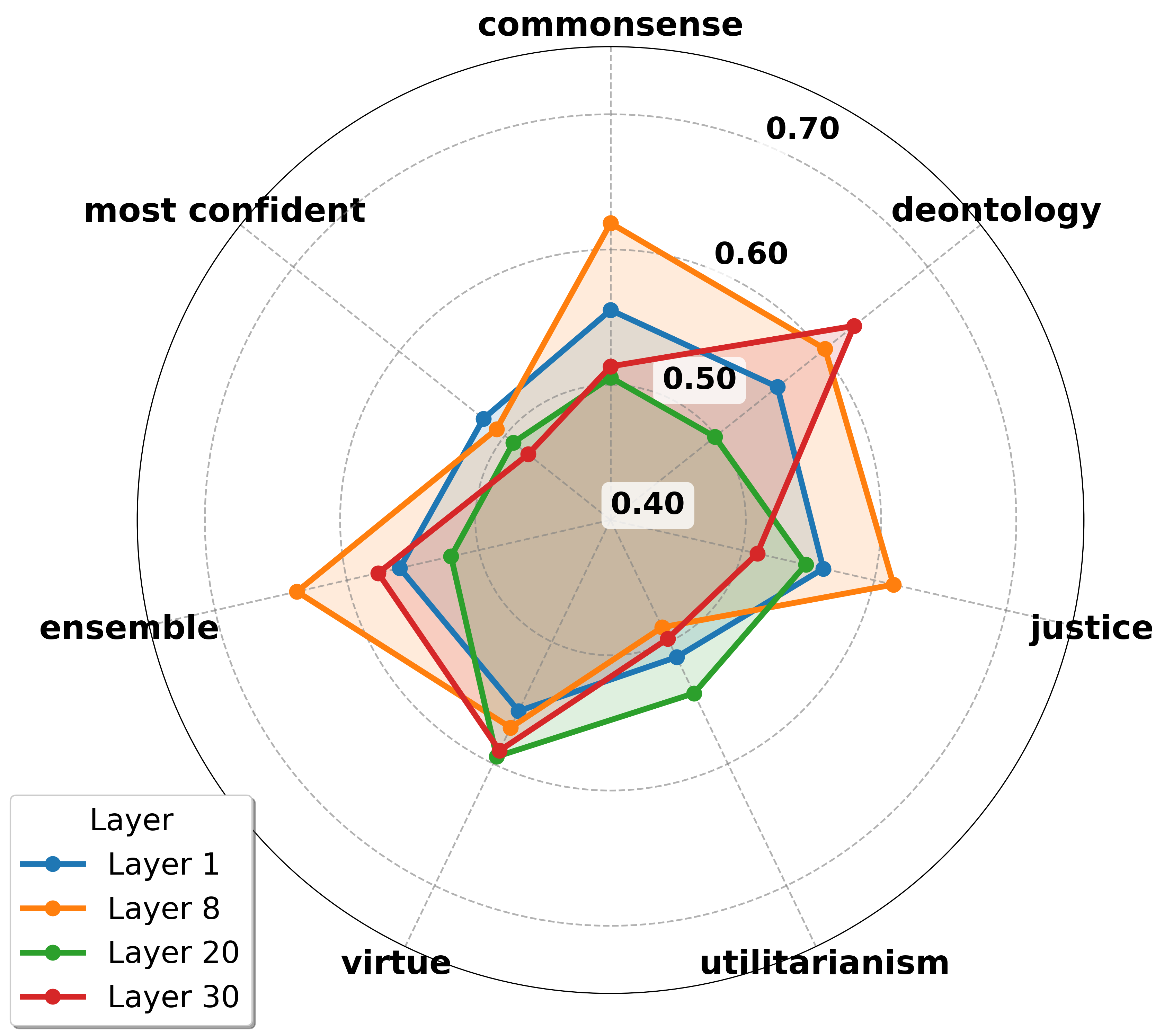}
    \caption{Mistral-7B-Instruct}
    \label{fig:mistral_spider}
  \end{subfigure}
  
  \vspace{0.3cm}
  
  \begin{subfigure}[t]{0.85\linewidth}
    \centering
    \includegraphics[width=\linewidth]{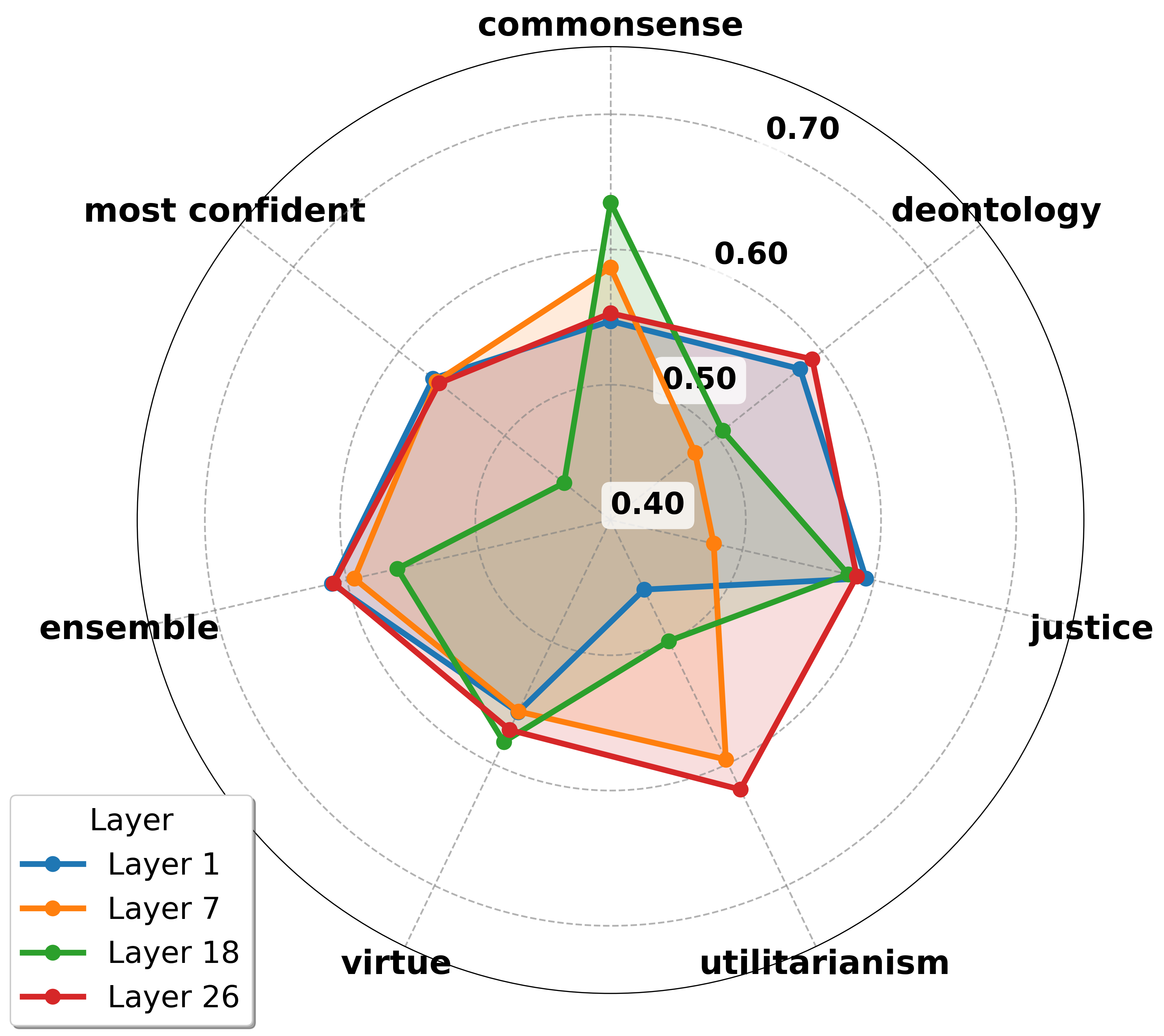}
    \caption{Qwen-2.5-7B-Instruct}
    \label{fig:qwen7b_spider}
  \end{subfigure}
  
  \vspace{0.3cm}
  
  \begin{subfigure}[t]{0.85\linewidth}
    \centering
    \includegraphics[width=\linewidth]{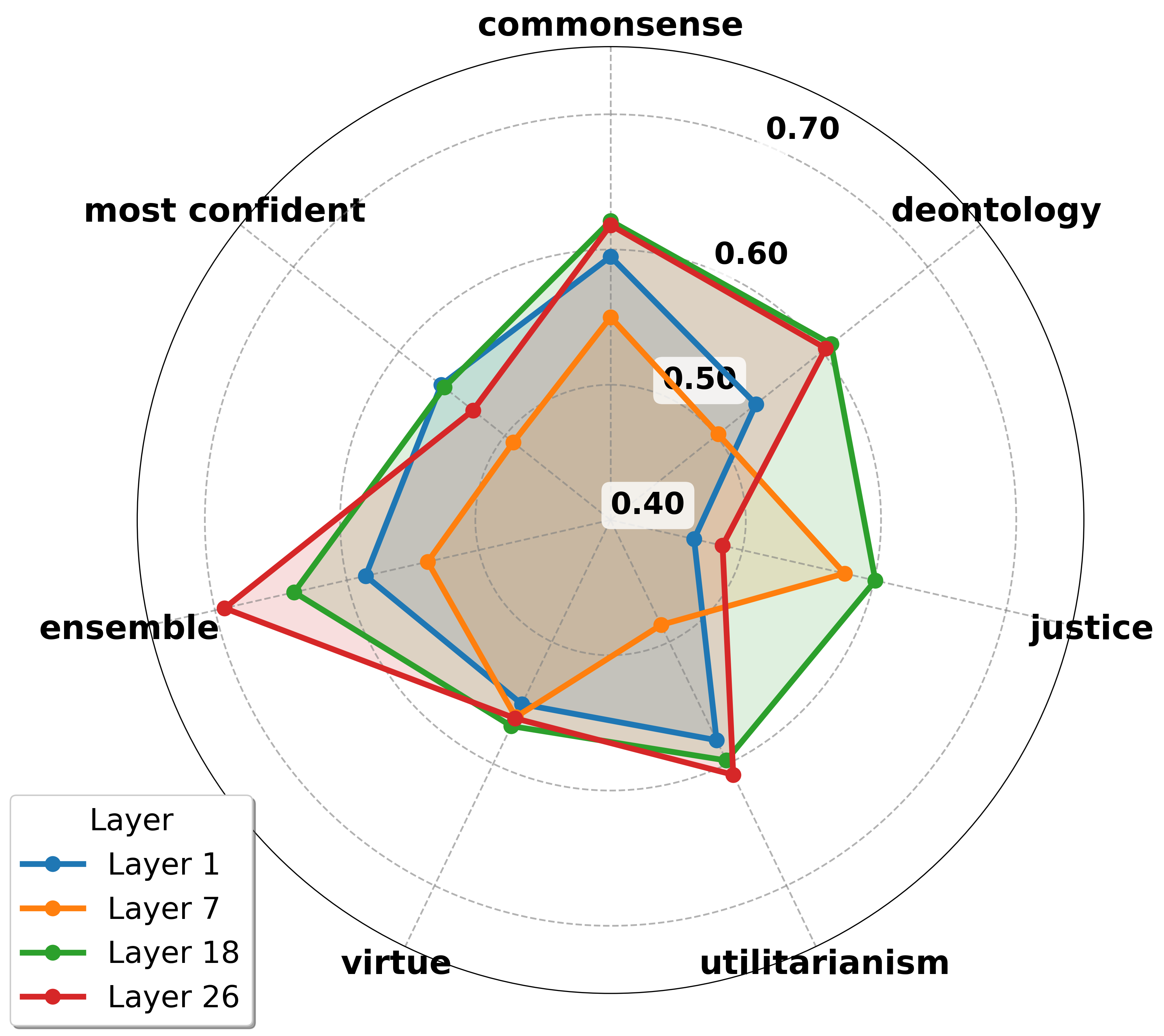}
    \caption{DeepSeek-R1-Distill-7B}
    \label{fig:deepseek_spider}
  \end{subfigure}
  
  \caption{Cross-domain generalization across 7B models at four network depths. Despite architectural differences and specialized training, all exhibit similar inflection points around 65\% depth.}
  \label{fig:7b_spider_comparison}
\end{figure}

Gemma-3-4B exhibits the generalization-specialization trade-off with compressed dynamics: generalization peaks at 65\% network depth (layer 16) before contracting toward specialization. The narrow optimal band for multi-framework reasoning (roughly 60--70\% depth) identifies constrained alignment leverage: early layers lack coherent ethical encoding, whereas deeper ones are already specialized.

\subsection{Medium-Scale Models (7B Parameters)}

Figure~\ref{fig:7b_spider_comparison} reveals consistent convergence in generalization dynamics across distinct 7B architectures. Despite different training objectives—standard instruction tuning (Mistral, Qwen) versus reasoning distillation (DeepSeek-R1)—all models exhibit peak cross-framework generalization at approximately 60--65\% network depth, with subsequent specialization toward final layers. The consistency of this inflection point across training paradigms suggests that the generalization-specialization trade-off emerges from shared capacity constraints rather than optimization procedures.

\begin{figure}[!t]
  \centering
  \begin{subfigure}[t]{0.85\linewidth}
    \centering
    \includegraphics[width=\linewidth]{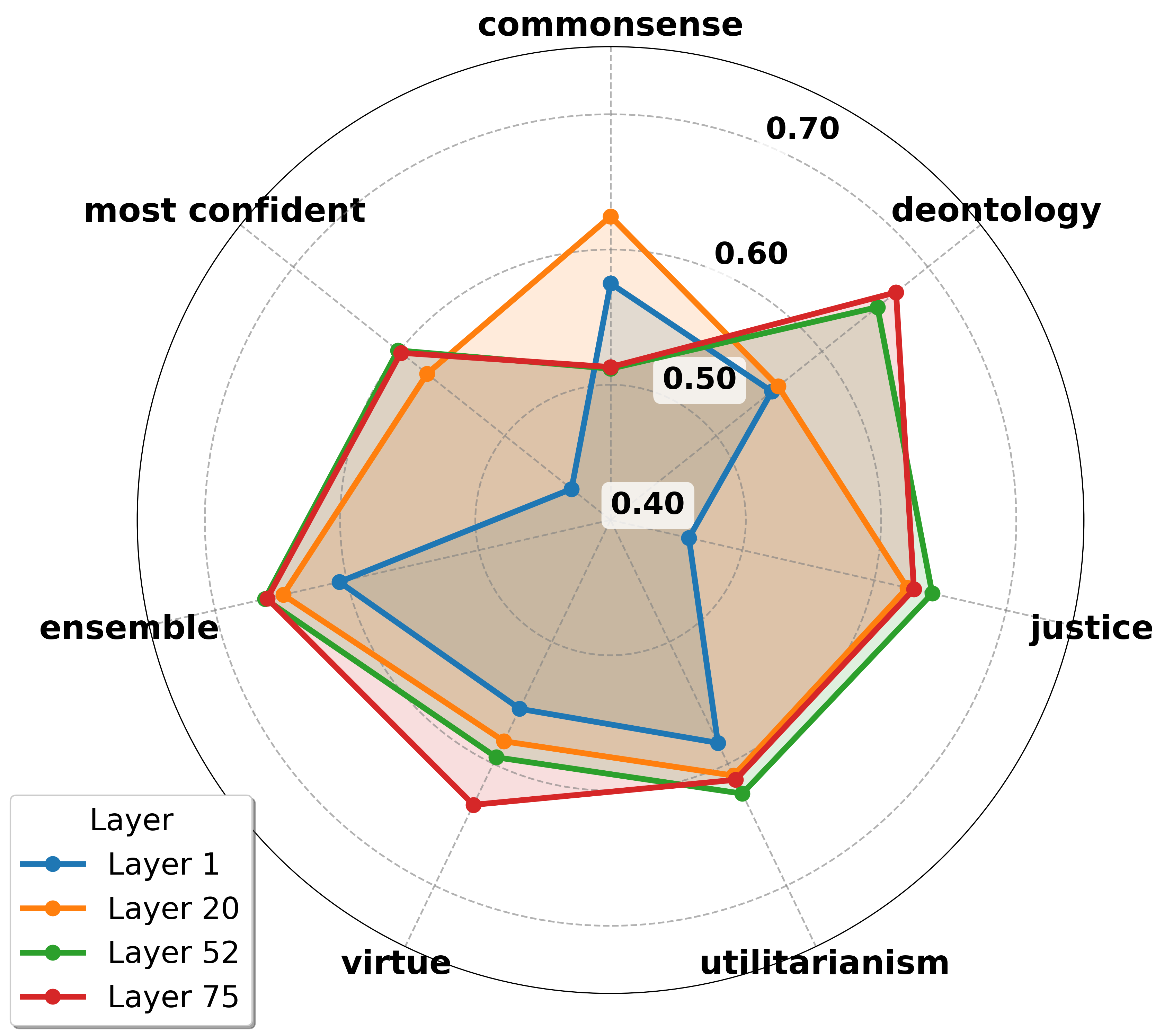}
    \caption{Llama-3.3-70B-Instruct}
    \label{fig:llama70b_spider}
  \end{subfigure}
  
  \vspace{0.3cm}
  
  \begin{subfigure}[t]{0.85\linewidth}
    \centering
    \includegraphics[width=\linewidth]{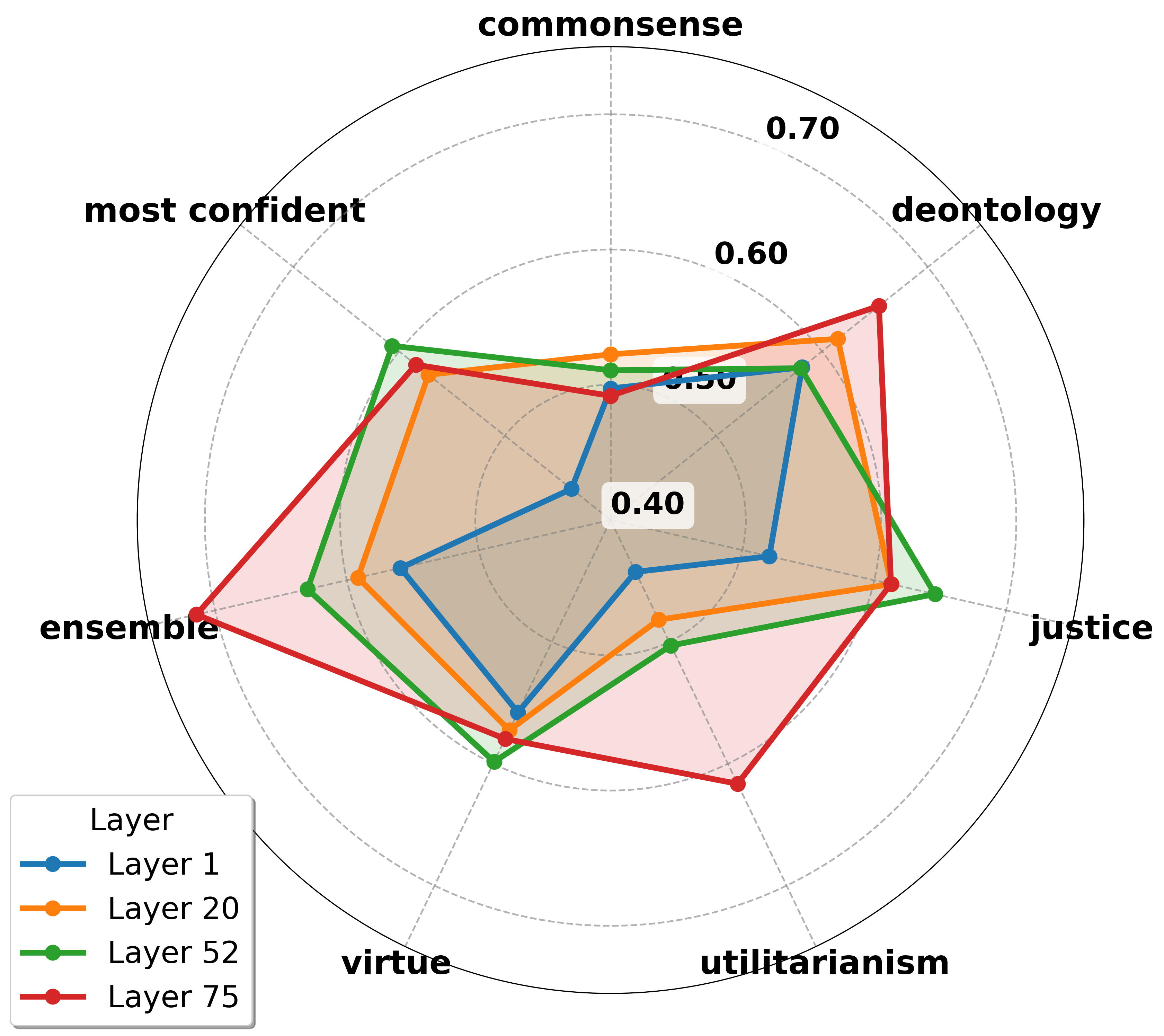}
    \caption{Qwen-2.5-72B-Instruct}
    \label{fig:qwen72b_spider}
  \end{subfigure}
  
  \caption{Cross-domain generalization for frontier-scale models at four network depths. Extended depth enables sustained high generalization across middle layers, with gradual rather than abrupt specialization transitions.}
  \label{fig:70b_spider_comparison}
\end{figure}

\begin{figure*}[!t]
\centering
\includegraphics[width=\linewidth]{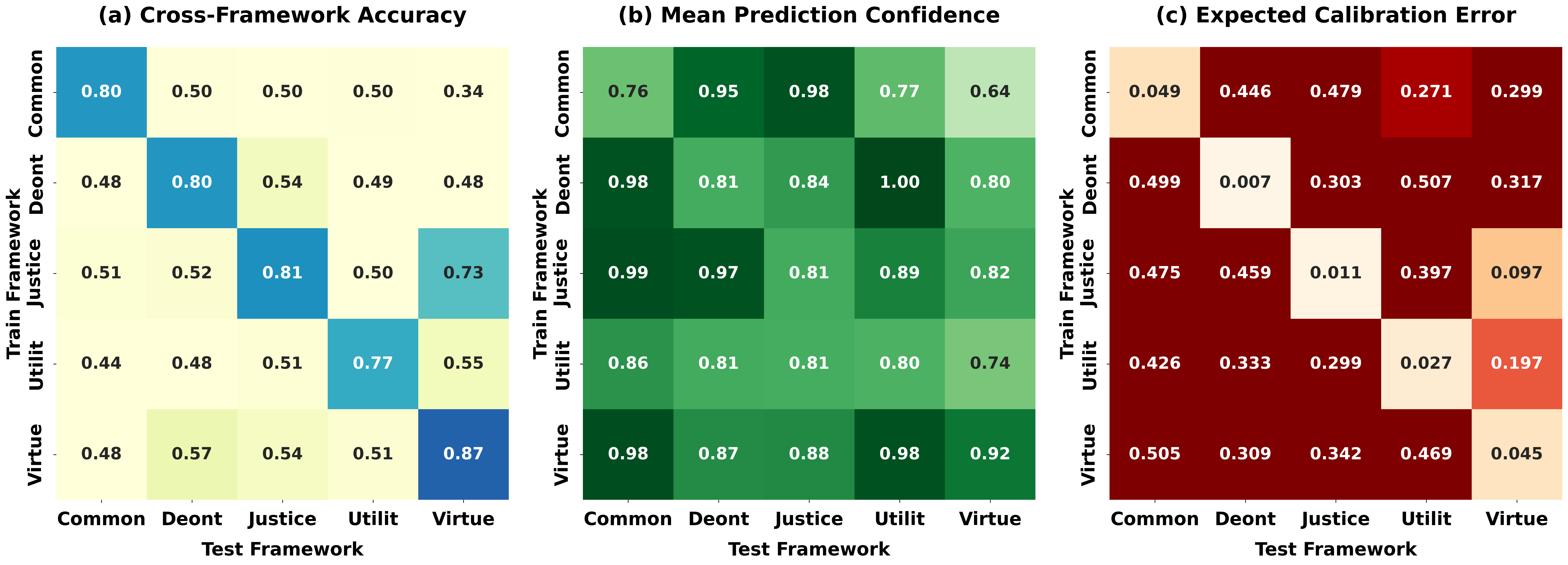}
\caption{Gemma-3-4B-Base shows that framework-specific representations emerge at minimal scale but with severe capacity-induced constraints. Catastrophic commonsense$\rightarrow$virtue transfer and extreme deontology$\rightarrow$utilitarianism miscalibration indicate the model learns discrete framework clusters without bridging representations.}
\label{fig:small_scale}
\end{figure*}

\subsection{Large-Scale Models (70B+ Parameters)}

Frontier models in Figure~\ref{fig:70b_spider_comparison} reveal how scale transforms generalization dynamics while preserving structural constraints. Both architectures achieve peak cross-domain performance at 65\% depth (layer 52), aligning with the proportional inflection point observed in 7B models. The distinguishing characteristic of frontier scale emerges in transition dynamics: from 25\% to 65\% depth, both models maintain robust cross-framework coherence with minimal degradation, contrasting with smaller models' rapid performance changes. This extended plateau—spanning 40\% of network depth—identifies a substantially wider candidate window for multi-framework alignment research.

The replication of patterns across nearly two orders of magnitude (4B to 72B) points to a consistent constraint: transformers cannot simultaneously maximize within-framework discrimination and cross-framework generalization beyond a threshold of representational maturity.

\section{Cross-Framework Transfer Across Model Scales}

Cross-framework entanglement patterns persist across diverse architectures and parameter counts, though manifestation varies by training methodology and scale.

\subsection{Scale-Invariant Failure Modes}

Three patterns emerge universally: strong within-framework discrimination, asymmetric cross-framework degradation, and miscalibrated confidence independent of accuracy. This consistency across 4B to 72B parameters indicates ethical framework entanglement reflects inherent constraints in neural language model representations rather than artifacts correctable through scaling.

\subsection{Small-Scale Models (4B Parameters)}

Gemma-3-4B-Base shows that framework-specific representations emerge even at minimal scale (Figure~\ref{fig:small_scale}). Within-framework discrimination succeeds despite limited capacity, but cross-framework transfer exposes clear architectural limitations. The catastrophic commonsense$\rightarrow$virtue failure suggests capacity constraints force the model into discrete representational clusters rather than continuous ethical spaces.

\subsection{Mid-Scale Models (7B Parameters)}

\begin{figure*}[!t]
\centering
\begin{subfigure}[t]{\linewidth}
    \centering
    \includegraphics[width=\linewidth]{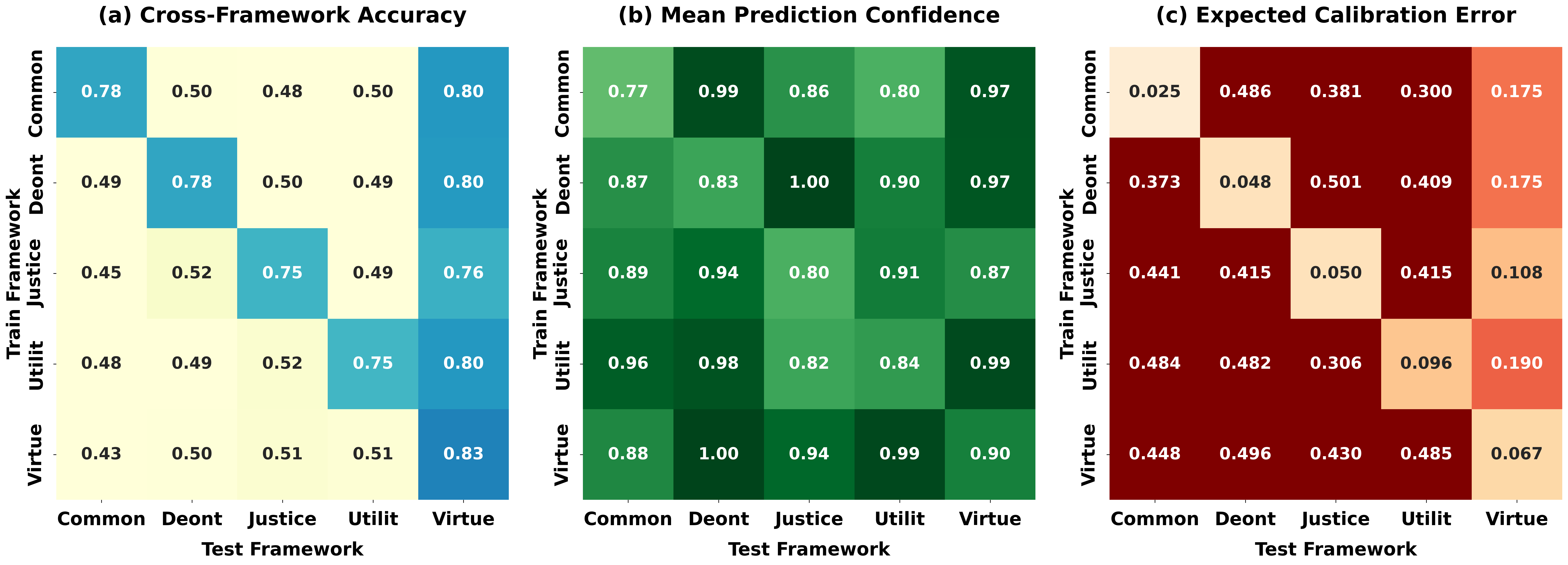}
    \caption{DeepSeek-R1-Distill-7B}
\end{subfigure}

\vspace{0.3cm}

\begin{subfigure}[t]{\linewidth}
    \centering
    \includegraphics[width=\linewidth]{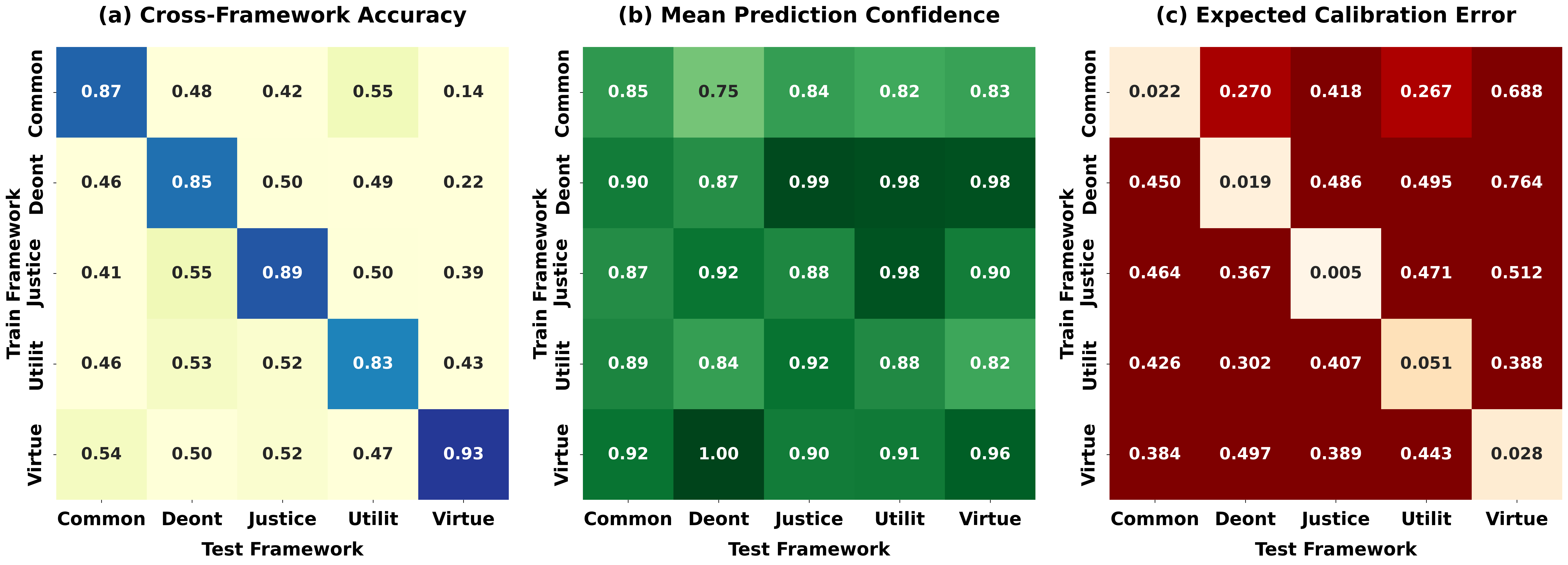}
    \caption{Mistral-7B-Instruct}
\end{subfigure}

\vspace{0.3cm}

\begin{subfigure}[t]{\linewidth}
    \centering
    \includegraphics[width=\linewidth]{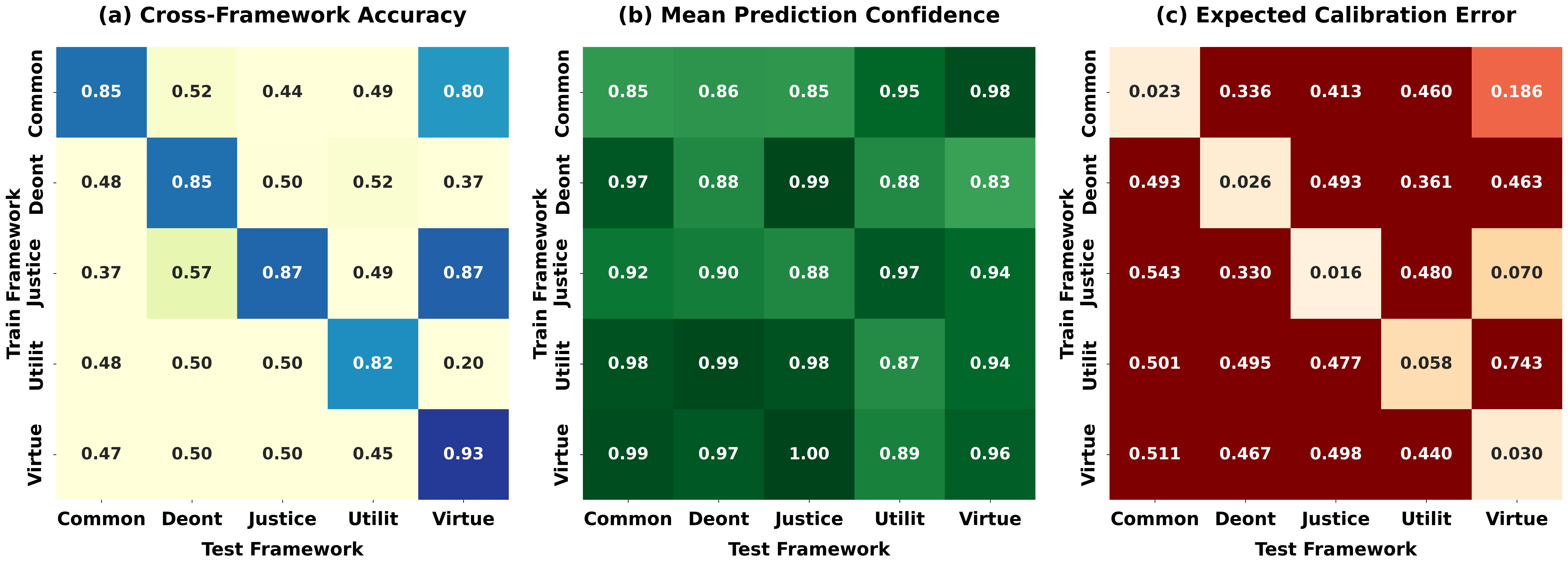}
    \caption{Qwen-2.5-7B-Instruct}
\end{subfigure}

\caption{Mid-scale models (7B) exhibit family-specific failure patterns. DeepSeek-R1 positions commonsense as uniquely problematic, potentially reflecting cultural-philosophical training biases. Mistral shows extreme virtue ethics isolation. Qwen demonstrates scale-invariant architectural biases—virtue isolation pattern identical to 72B variant.}
\label{fig:mid_scale}
\end{figure*}

The 7B parameter range reveals how training methodology shapes ethical representation geometry without eliminating entanglement (Figure~\ref{fig:mid_scale}).

DeepSeek-7B positions commonsense morality as uniquely problematic—probes trained on commonsense fail systematically across all frameworks. This asymmetry suggests DeepSeek's training emphasized commonsense as a distinct reasoning mode, potentially reflecting Chinese cultural philosophical traditions that conceptualize common sense differently from Western utilitarian-deontological frameworks. DeepSeek's divergent failure pattern provides evidence that our probing method is sensitive to cultural variation in moral reasoning rather than imposing uniform Western assumptions.

Mistral-7B-Instruct exhibits the most severe transfer asymmetries at this scale. Virtue ethics forms an isolated representational island—incoming transfers approach chance while outgoing transfers succeed moderately. This pattern suggests Mistral's instruction tuning data contained virtue ethics examples that were structurally distinct from other frameworks.

Qwen-2.5-7B-Instruct exhibits family-specific architectural biases that persist across scale. The virtue ethics isolation pattern nearly identical to Qwen-2.5-72B indicates that Qwen's training methodology creates systematic representational structure independent of capacity.

\subsection{Large-Scale Models (70B-72B Parameters)}

\begin{figure*}[!t]
\centering
\begin{subfigure}[t]{\linewidth}
    \centering
    \includegraphics[width=\linewidth]{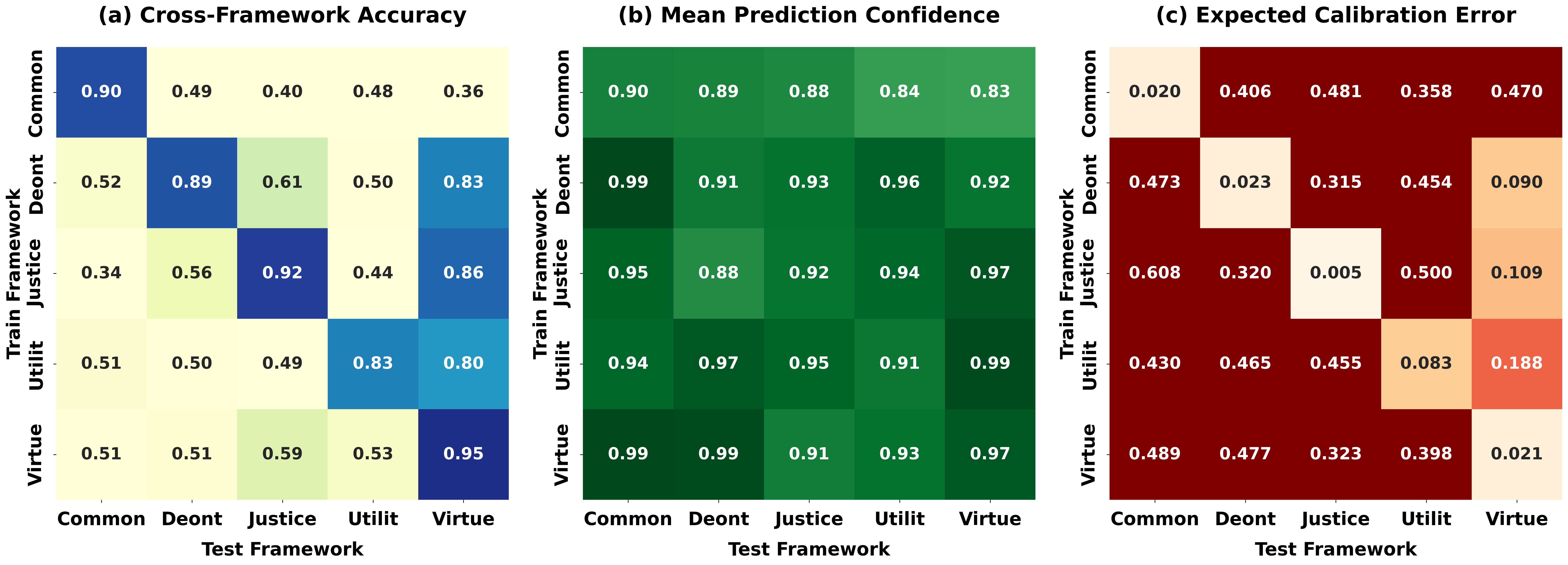}
    \caption{Llama-3.3-70B-Instruct}
\end{subfigure}

\vspace{0.3cm}

\begin{subfigure}[t]{\linewidth}
    \centering
    \includegraphics[width=\linewidth]{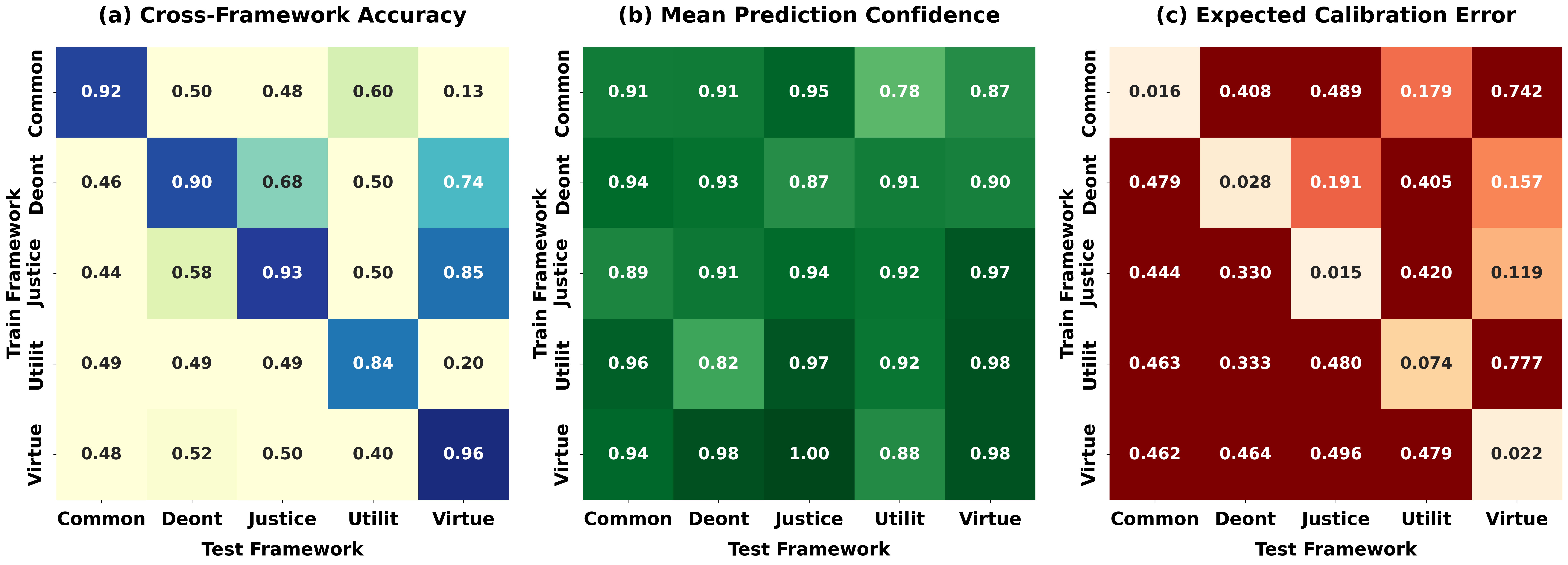}
    \caption{Qwen-2.5-72B-Instruct}
\end{subfigure}

\caption{Frontier-scale models show capacity limits on resolving entanglement. Llama-70B shows accuracy-calibration dissociation—improved within-framework performance but persistent severe out-of-domain miscalibration. Qwen-72B exhibits pronounced virtue isolation despite 72B parameters, preserving patterns from 7B.}
\label{fig:large_scale}
\end{figure*}

Llama-3.3-70B-Instruct achieves the strongest overall performance but preserves core entanglement patterns (Figure~\ref{fig:large_scale}). The key observation: accuracy and calibration dissociate at scale. Llama achieves higher domain-specific accuracy while maintaining severe out-of-domain miscalibration, indicating standard training objectives optimize for confident correct responses without teaching distribution shift detection.

Qwen-2.5-72B-Instruct shows that extreme scale cannot overcome architectural or training methodology biases. Despite 72B parameters, the model exhibits the most pronounced virtue ethics isolation across all evaluations. The preservation of this pattern from Qwen-2.5-7B indicates systematic rather than capacity-limited failures.

\subsection{Cross-Scale Insights}

Three architectural insights emerge from cross-scale analysis. First, model families display consistent failure modes independent of scale—Qwen's virtue isolation, Mistral's strong asymmetries, and Llama's balanced yet entangled structure—indicating that training methodology shapes which frameworks interfere, while architecture governs interference magnitude. Second, instruction tuning amplifies within-framework confidence without improving cross-framework calibration, suggesting that alignment training favors decisiveness over calibrated uncertainty. Third, scaling enhances accuracy within frameworks but fails to mitigate inter-framework interference or enable detection of out-of-distribution ethical contexts.

The persistence of high-confidence miscalibration across scales and architectures is arguably the most significant result: models show no reliable uncertainty signal when applying ethical frameworks to out-of-distribution scenarios. This failure mode appears invariant to capacity, architecture, and training, reflecting a structural limitation in how current language models represent normative uncertainty.

\section{Probe Conflict and Behavioral Inconsistency in Llama-3.3-70B-Instruct}
\label{appendix:llama_conflict}

\begin{figure}[!t]
    \centering
    \begin{subfigure}[t]{\linewidth}
        \centering
        \includegraphics[width=\linewidth]{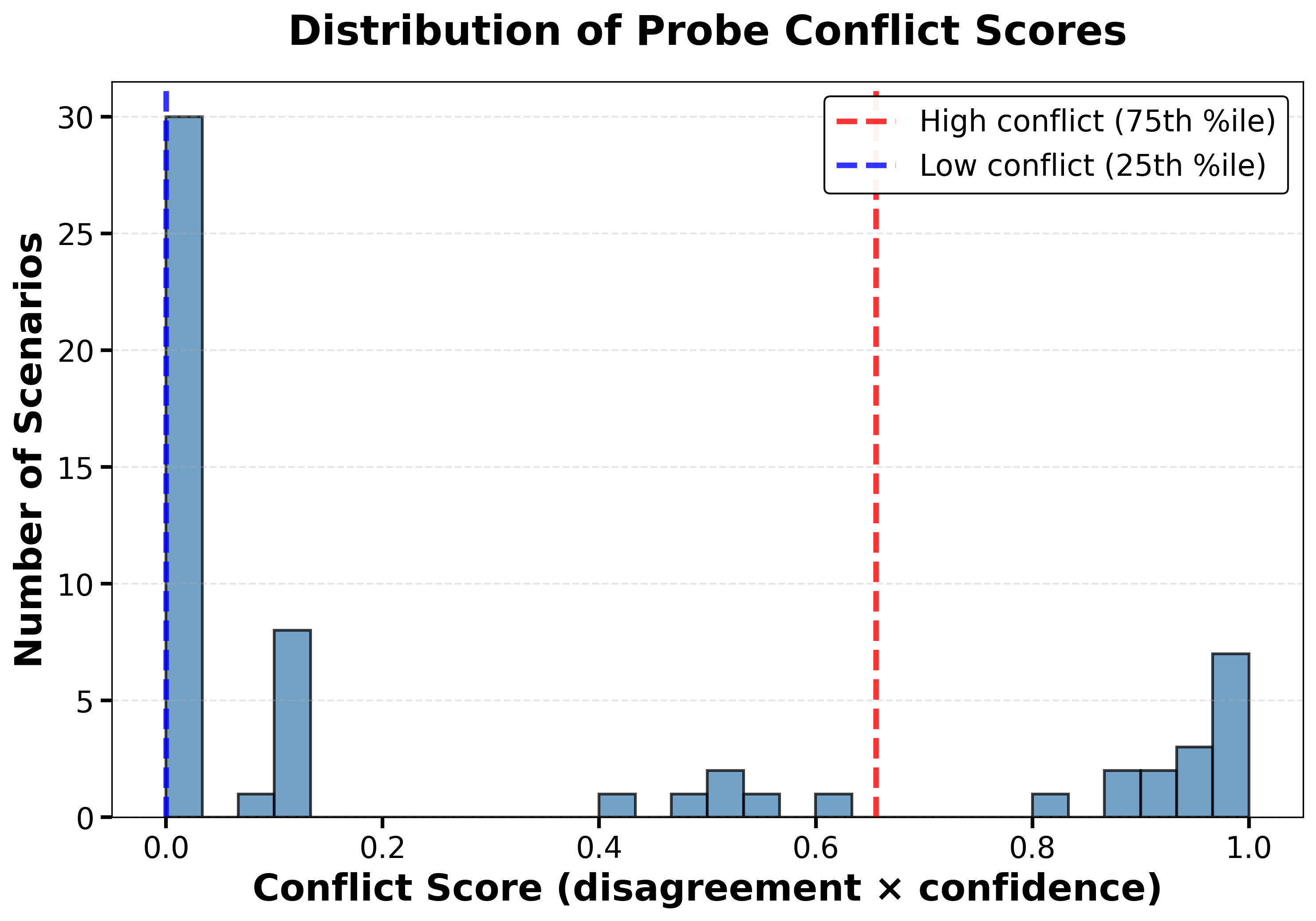}
        \caption{Distribution of conflict scores. The right-skewed distribution with concentration at zero contrasts with Mistral-7B's bimodal pattern, suggesting greater representational overlap in larger aligned models.}
        \label{fig:llama_conflict_dist}
    \end{subfigure}
    
    \vspace{0.5em}
    
    \begin{subfigure}[t]{\linewidth}
        \centering
        \includegraphics[width=\linewidth]{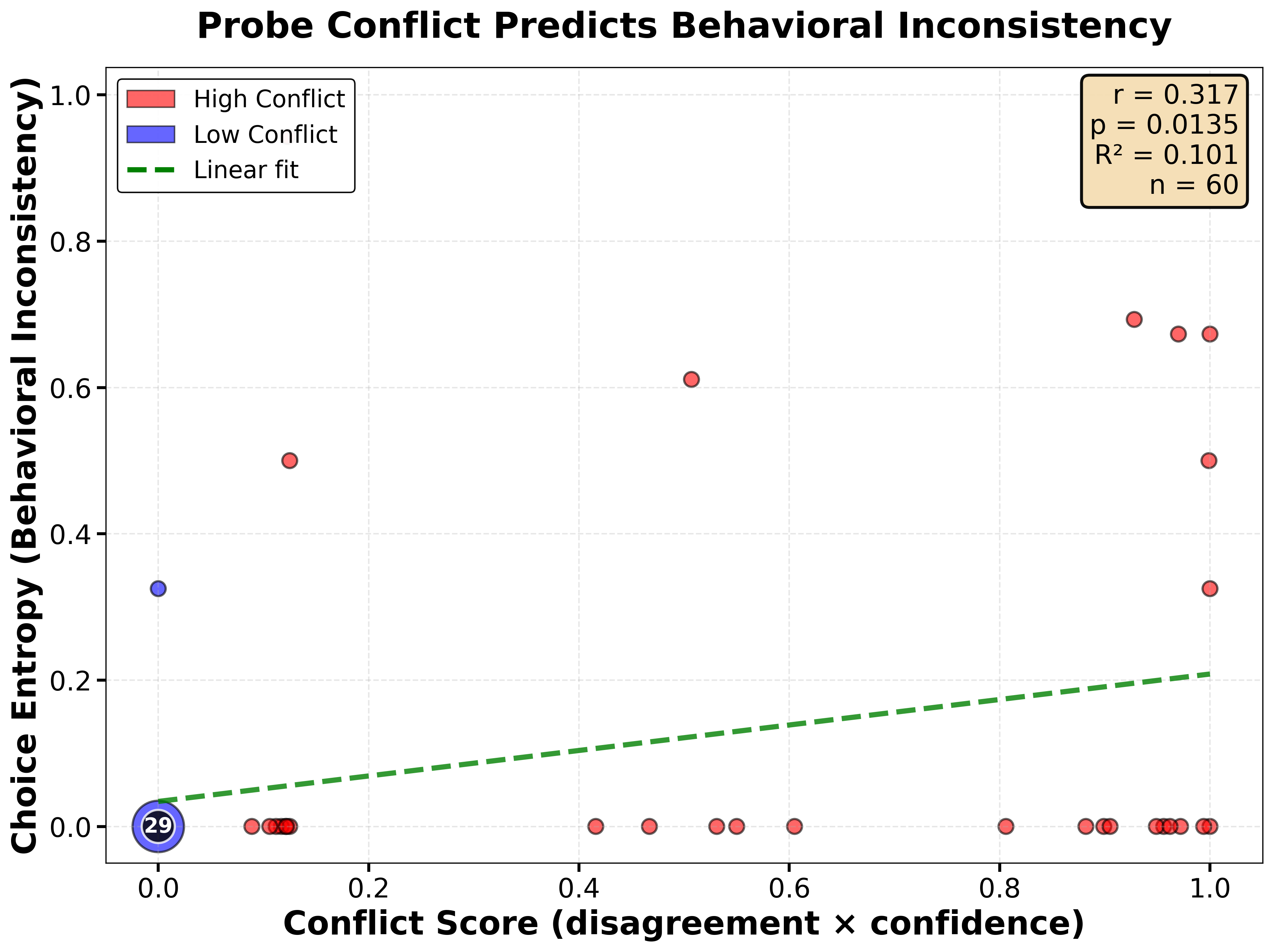}
        \caption{Relationship between conflict scores and choice entropy ($T=1.2$). Despite the skewed distribution, scenarios with higher conflict correlate with greater behavioral inconsistency ($r = 0.317$, $p = 0.014$).}
        \label{fig:llama_conflict_scatter}
    \end{subfigure}
    \caption{Probe conflict correlates with behavioral inconsistency in Llama-3.3-70B-Instruct.}
    \label{fig:llama_conflict}
\end{figure}

To examine how the conflict-entropy relationship manifests at frontier scale, we conducted the same analysis on Llama-3.3-70B-Instruct (Figure~\ref{fig:llama_conflict}).

Unlike Mistral-7B's bimodal distribution, Llama-3.3-70B exhibits a heavily right-skewed conflict profile: nearly half of test scenarios show zero conflict, with a long tail of high-conflict cases. This concentration at zero suggests that despite being trained on distinct ethical principles, the deontological and utilitarian probes predominantly agree in the larger model. The pattern indicates potential representational overlap—framework-specific signatures may partially merge toward a simpler acceptable/unacceptable distinction under heavy alignment training.

To assess whether this concentration at zero reflects a genuine property of the representations rather than an artifact of probe training, we validated the stability of the conflict metric via bootstrapping. We retrained probes using 5 different random seeds on resampled training data and compared the resulting conflict scores. The continuous conflict signal showed strong stability (Pearson $r \approx 0.76$ across training folds), indicating that the probes detect a consistent geometric direction of conflict in activation space. However, the binary ``zero-conflict'' set (bottom 25\%) showed lower overlap across folds (IoU $\approx 0.29$), suggesting that while the relative ranking of scenarios by conflict is robust, the specific threshold separating zero-conflict from low-conflict scenarios is sensitive to sampling noise.

Despite the skewed distribution, probe conflict correlated with behavioral inconsistency. High-conflict scenarios showed greater choice entropy than low-conflict ones ($p = 0.014$, Cohen's $d = 0.54$). The continuous correlation was modest ($r = 0.317$), weaker than Mistral-7B's $r = 0.36$. This attenuation may reflect alignment training suppressing the behavioral expression of representational conflicts—the underlying tensions exist but manifest less transparently in outputs.

The large marker at $(0,0)$ in Figure~\ref{fig:llama_conflict_scatter} represents 29 scenarios where probes agreed and models produced consistent responses, illustrating the predominance of low-conflict, consistent cases. Red markers (high-conflict scenarios) nonetheless show elevated entropy, confirming that when representational disagreement does occur, it propagates to behavioral inconsistency even in heavily aligned frontier models.

This pattern carries implications for probe-based auditing at scale. While smaller models like Mistral-7B offer clearer diagnostic windows due to preserved framework differentiation, the conflict-entropy relationship remains significant in Llama-3.3-70B—suggesting that the conflict-entropy correlation persists at scale, though its diagnostic utility remains to be validated after controlling for difficulty confounds.

\end{document}